\documentclass[twoside,11pt]{article}
\usepackage{jair, theapa, rawfonts}

\usepackage{todonotes}
\usepackage[nolist]{acronym}
\usepackage{amsmath}
\usepackage{amsfonts}
\usepackage{multirow}
\usepackage{subfig}
\usepackage{graphicx}
\usepackage{url}
\usepackage{booktabs}

\newtheorem{idefinition}{Definition}

\ShortHeadings{Compass-aligned Distributional Embeddings for Studying Semantic Differences across Corpora}
{Bianchi, Di Carlo, Nicoli \& Palmonari}
\firstpageno{1}

\begin{document}

\title{Compass-aligned Distributional Embeddings for Studying Semantic Differences across Corpora}

\author{\name Federico Bianchi\thanks{Part of this work was carried out when Federico Bianchi was a PhD student at University of Milano-Bicocca} \email f.bianchi@unibocconi.it \\
      \addr Bocconi University, Milan, Italy
        \AND
        \name Valerio Di Carlo \email valerio.dicarlo@bupsolutions.com \\
      \addr BUP Solution, Rome, Italy
      \AND
        \name Paolo Nicoli \email p.nicoli@campus.unimib.it \\
      \addr University of Milan-Bicocca, Milan, Italy
      \AND
        \name Matteo Palmonari \email matteo.palmonari@unimib.it \\
      \addr University of Milan-Bicocca, Milan, Italy
        }



\maketitle

\begin{abstract}
Word2vec is one of the most used algorithms to generate word embeddings because of a good mix of efficiency, quality of the generated representations and cognitive grounding. However, word meaning is not static and depends on the context in which words are used. Differences in word meaning that depends on time, location, topic, and other factors, can be studied by analyzing embeddings generated from different corpora in collections that are representative of these factors. For example, language evolution can be studied using a collection of news articles published in different time periods. In this paper, we present a general framework to support cross-corpora language studies with word embeddings, where embeddings generated from different corpora can be compared to find correspondences and differences in meaning across the corpora. \ac{atmodel} is the core component of our framework and solves the key problem of aligning the embeddings generated from different corpora. In particular, we focus on providing solid evidence about the effectiveness, generality, and robustness of CADE. 
To this end, we conduct quantitative and qualitative experiments in different domains, from temporal word embeddings to language localization and topical analysis. The results of our experiments suggest that CADE achieves state-of-the-art or superior performance on tasks where several competing approaches are available, yet providing a general method that can be used in a variety of domains. Finally, our experiments shed light on the conditions under which the alignment is reliable, which substantially depends on the degree of cross-corpora vocabulary overlap. 
\end{abstract}

\begin{acronym}
\acro{atmodel}[CADE]{Compass Aligned Distributional Embeddings}
\acro{twem}[TWEM]{Temporal Word Embedding Model}
\acro{twe}[TWE]{Temporal Word Embedding}
\acro{datas}[NAC-S]{News Article Corpus Small}
\acro{datab}[NAC-L]{News Article Corpus Large}
\acro{dataml}[MLPC]{Machine Learning Papers Corpus}
\acro{tests}[T1]{Testset1}
\acro{testb}[T2]{Testset2}
\acro{wem}[WEM]{Word Embedding Model}
\acro{twa}[TWA]{Temporal Word Analogy}
\acro{staticness}[STAT]{Staticness}
\end{acronym}
\tableofcontents

\section{Introduction}
\label{Introduction}

First introduced in the fifties, the distributional hypothesis~\cite{Harris1985DistributionalStructure,Firth1957A1930-1955} laid the basis for a different point of view on word meaning. The distributional hypothesis advocates for a usage-based perspective on language. In brief, \textit{the meaning of a word is function of the contexts in which it appears}: words like ``cat'' and ``dog'' are expected to appear in similar contexts and thus be more similar than words like ``cat'' and ``frogman'', which are expected to appear in different contexts. Theories and models that account for the meaning of words (or other language expressions - but in the following, we will focus on words only) and are inspired by this hypothesis are usually considered part of \textit{distributional semantics}.   

Inspired by the distributional semantics, researchers have developed models where words are represented by $n$-dimensional dense vectors that are derived from the usage of words in some text corpus using different approaches, from count-based methods to neural networks~\cite{Baroni2014DontVectors,Lenci2017DistributionalMeaning}. These vector-based representations are also referred to as \textit{distributed representations}, \textit{word embeddings}, or \textit{distributional representations} when the method used to generate the vectors is more grounded in the distributional hypothesis. Under distributional semantics, vectors representing similar words are expected to be close in the vector space, with similarity and distance functions in the vector space interpreted as semantic similarity or distance measures. Word2vec~\cite{mikolov2013exploiting} is one of the most used algorithms to generate word embeddings because of a good mix of efficiency, quality of the generated representations and ties with the distributional hypothesis. Previous work has discussed the cognitive grounding of representations generated with word2vec from representative corpora~\cite{Lenci2008DistributionalResearch}, which makes them quite appealing for language studies, and, for example, they have been used to analyze biases in language~\cite{caliskan2017semantics}.    

\begin{figure*}[h!]
\centering
\includegraphics[width=1\textwidth]{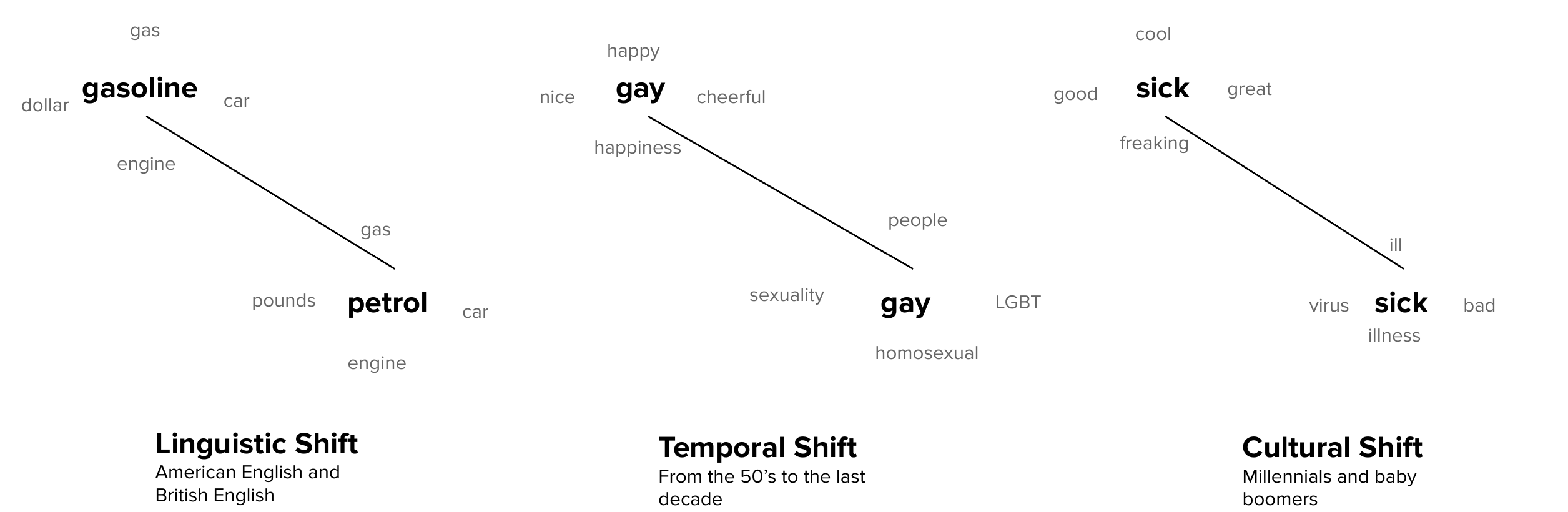}
\caption{Examples of meaning shifts.}
\label{fig:meaning:shift} 
\end{figure*}

However, word meaning is neither constant nor universal. For example, the usage of some words has changed significantly across time, e.g., from the 50s to today. Think about the \textit{core} meaning of ``gay'', which has shifted from the 50s to today as a consequence of being used predominantly as a synonym of ``joyous'' (in the 50s) vs. as an indication of sexual orientation (today)~\cite{Hamilton2016DiachronicChange}. Analogously, the core meaning of ``amazon'' in news articles has shifted from referring to a forest (until the 90s) to identifying a company (today), as a consequence to the change of contexts in which it is predominantly used~\cite{Yao2017DiscoveryLearning}\footnote{The reference to core meaning is borrowed from~\cite{hamilton2016cultural} and considers the known problem of polysemous words; in this paper, we focus on word-level embeddings where a token is associated with a unique vector representing its core meaning. However, approaches proposed to solve or mitigate the disambiguation problem for polysemous words within word-level embeddings~\cite{iacobacci2015sensembed} can be used in combination with the approach proposed in this paper; otherwise, the contextual word embeddings approaches, which natively solves the word ambiguity problem, also limitations that make them more difficult to be used to study meaning shift and related patterns; for further insights into this discussion, we refer to Section~\ref{relatedwork}}. The meaning of words can change also depending on the location, as we can observe in language localization, e.g., when we compare the core meaning of ``flat'' in American-English and British-English, where the word is mainly used to refer to a shape of surfaces (in American-English) vs. to apartments (in British-English). Word embeddings have in fact been proposed as valuable models to support language studies, mainly in the context of the analysis of language evolution~\cite{Hamilton2016DiachronicChange,Yao2017DiscoveryLearning,dicarlo2019}, but also in the context of location-based longitudinal analyses~\cite{Bamman2014DistributedLanguage}. While the phenomenon under attention may be framed differently, e.g., as meaning shift vs. cultural drift~\cite{hamilton2016cultural}, all these studies have in common that differences in meaning, i.e., \textit{semantic differences}, are studied by comparing word usage in different corpora. Figure~\ref{fig:meaning:shift} shows some examples of semantic differences that can be framed as different kinds of meaning shift and could be studies using cross-corpora word-level semantic comparisons. 

\begin{figure*}[h!]
\centering
\includegraphics[width=0.50\textwidth]{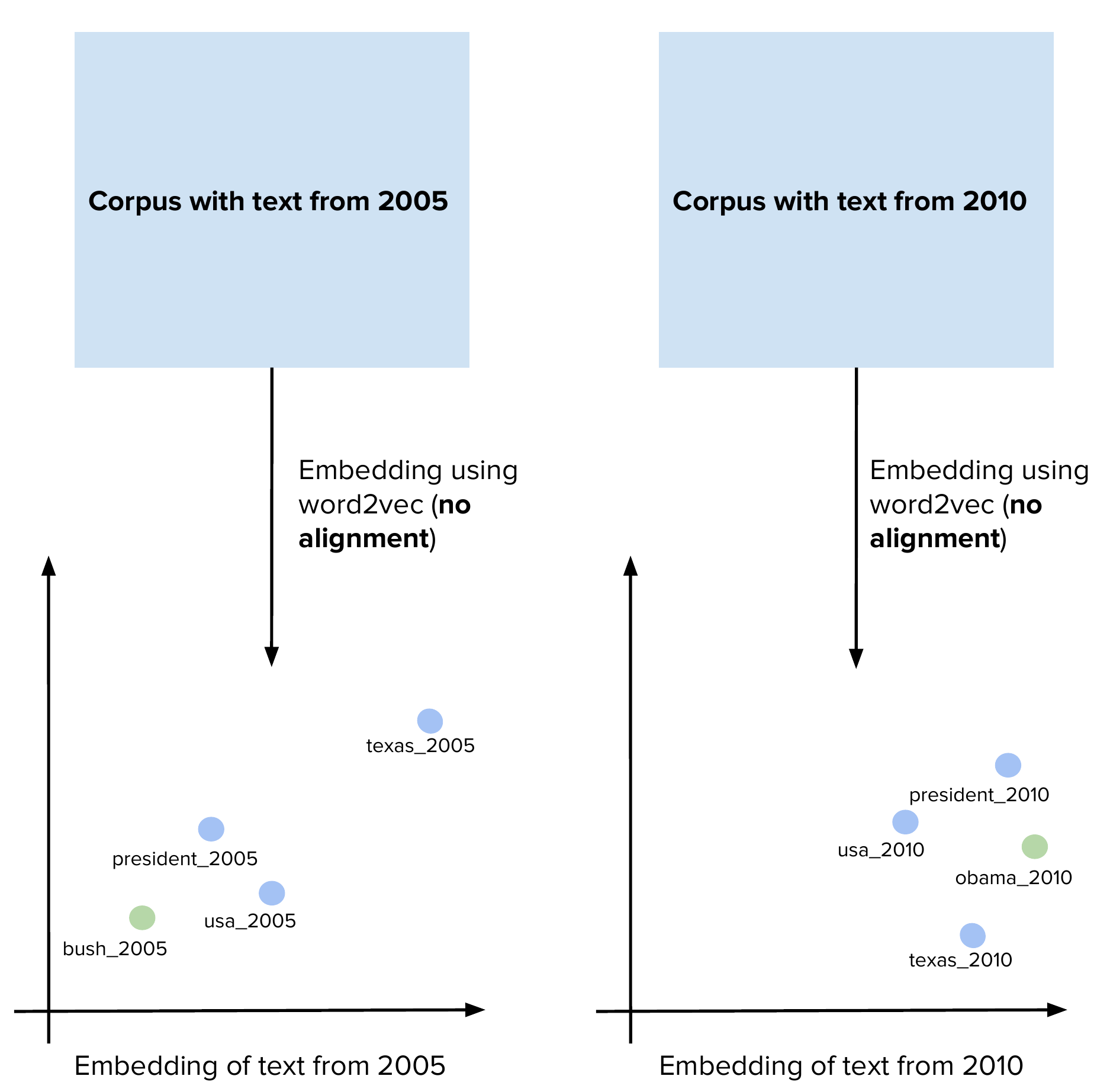}
\caption{Word embeddings generated from two different corpora are not directly comparable. The word \textit{president}, for example, occupies to completely different positions in the vector spaces.}
\label{fig:cade:alingment:problem} 
\end{figure*}

A simple way to study meaning shift and other semantic differences in one language would be to generate different word embeddings with word2vec having equal vector dimension for each corpus in a collection or, analogously, for each \textit{slice} that results from splitting a collection according to some criterion. Criteria used to split a collection may be different, such as time of publication (e.g., 2001 articles, 2002 articles, etc.), language localization (e.g., American-English and British-English), source of publication (e.g., The New York Times, The Washington Post, The Guardian, etc.), topic (e.g., Science, Science Fiction, etc.), and so on. However,  the low-level stochasticity of neural networks used to train word2vec do not allow to generate comparable representations, which means that similarity computed between vectors generated from different corpora (or slices) would not be meaningful from a semantic point of view. For example, training the word2vec algorithm two times on the same slice generates different vectors for the same words, because the ``coordinate system'' on which the vectors are embedded would be different. Another example is depicted in Figure~\ref{fig:cade:alingment:problem}: training two corpora of articles published respectively in 2005 and 2010 would assign different positions to a word like ``president'', whose meaning has arguably not changed across the two time periods.    
A close analogy would be to ask two cartographers to draw a map of USA during different periods, without giving either of them a compass: the two maps would be similar, although one will be rotated by an unknown angle with respect to the other~\cite{Smith2017ODictionary}. To support meaningful comparison among embeddings generated for different corpora (or, analogously, slices of one collection), these embeddings need to be \textit{aligned} to make sure that vectors associated with words whose meaning is not expected to change across corpora are stable across the embeddings associated with each corpus. 
In other words, semantic comparison across corpora requires a reliable solution to the \textit{alignment problem}.       


Most of state of the art approaches proposing solutions to the alignment problem, have been targeted to temporal meaning shift and focused on aligning embeddings generated from different temporal slices~\cite{kutuzov-etal-2018-diachronic,Kulkarni2015StatisticallyChange,Szymanski2017TemporalEmbeddings,rudolph2016exponential,Yao2017DiscoveryLearning} or different languages~\cite{conneau2017word}; little work has explored alignment in the context of language localization~\cite{Bamman2014DistributedLanguage,galliani2019}. 

In this paper, we present an approach to support the alignment of word embeddings generated from different corpora and a semantic comparison framework that exploits aligned embeddings to support cross-corpora semantic comparisons. The core of our approach is an \textit{unsupervised} method that we refer to as \acf{atmodel}, which supports the implicit alignment between corpus-specific embeddings. In short, the approach leverages the two weights matrices used to train the Continuous Bag of Words (CBOW) model~\cite{Mikolov2013DistributedCompositionality} in word2vec: one matrix is updated with the input from a specific corpus, while a second matrix is frozen after being previously trained over the whole collection. The weights in the first matrix provide corpus-specific embeddings, the ones in the second matrix - the compass - derive from collection-wide word usage. The approach has been previously applied to align temporal word embeddings~\cite{dicarlo2019}. In experiments with temporal word embeddings, the approach has achieved state-of-the-art or superior performance if compared to other approaches on both small and large text collections, despite its simplicity and independence from time-specific assumptions, e.g., a linear order between corpora. The latter feature, as well as its efficiency, made it natural to generalize the proposed approach as part of a semantic comparison framework that can be used, under certain conditions, with any set of corpora. Our framework is available online\footnote{\url{https://github.com/vinid/cade}}.

\subsection{Summary of Contributions}

The overall goal of the proposed framework is to support language-based studies by domain experts, which explains why simplicity and efficiency are two desired properties. In addition to framing the alignment method developed in our previous work~\cite{dicarlo2019} into a framework for cross-corpora semantic comparison, in this paper we focus in particular on providing solid evidence about 1) the performance of the approach when compared to other alignment strategies, 2) its cross-domain generalization potential, and 3) its robustness, including a characterization of the conditions under which the alignment is more successful (as a function of cross-corpora vocabulary overlap). 

To provide such evidence, quantitative experiments are conducted in domains where CADE can be compared to previous work because consolidated test data and methodologies are available, e.g., in domains like temporal shift and language localization. To discuss its potential as a general framework to support semantic comparisons, we also discuss more qualitative experiments in domains where hard test data are not available, e.g., in the context of topic-wise comparisons.

As a summary, CADE provides, to the best of our knowledge, the first approach that can be used to generate comparable distributional models of words independent from the kind of comparison, yet achieving state-of-the-art results in contexts like temporal comparison where several specific approaches have been provided.   

The paper is structured as follows:
In Section~\ref{relatedwork} we describe the related work focusing on approaches that account for temporal alignment. In Section~\ref{sec:compass} we introduce CADE, describing its main properties and characteristics. Sections~\ref{sec:termoral:word:embeddings},\ref{sec:experiments:generalization},\ref{sec:experiments:robustness} describe respectively the experiments on temporal alignment, on language localization and on the robustness of our method. Eventually, we conclude the paper in Section~\ref{sec:conclusions}, summarizing what we have presented.

\section{Related Work}\label{relatedwork}

In this Section, we first give a high-level overview of the topic of semantic comparison and meaning shift. Then the focus changes to how, in the most modern research in this field, the comparison between different collections using distributional embeddings has been tackled: the two main categories of approaches are temporal word embeddings and multi-lingual word embeddings. The Table~\ref{tab:sota:comparison} shows the approaches modern research has taken that consider the problem of meaning shift by also underlining if they have been applied to multiple collections or only to a single collection.

\begin{table}[]
    \centering
    \small
    \begin{tabular}{cccc} \toprule
         Approach & Multiple Corpora & Shift Type   \\ \midrule
         \cite{Kulkarni2015StatisticallyChange} & Yes & Temporal  \\
         \cite{Hamilton2016DiachronicChange} & Yes & Temporal  \\
         \cite{Szymanski2017TemporalEmbeddings} & Yes & Temporal  \\ \midrule
         \cite{Yao2017DiscoveryLearning} & Yes & Temporal  \\
         \cite{Tripodi2019antisemitic} & Yes & Temporal  \\
         \cite{garg2018word} & Yes & Temporal   \\
         \cite{Bamler2017DynamicEmbeddings} & Yes & Temporal   \\ 
         \cite{Bamman2014DistributedLanguage} & Yes & Temporal   \\ \midrule
         \cite{caliskan2017semantics} & No & Cultural   \\
         \cite{galliani2019} & Yes & Cultural   \\ \bottomrule
    \end{tabular}
    \caption{Different models in the state-of-the-art}
    \label{tab:sota:comparison}
\end{table}{}

\subsection{Overview on the State of the Art}

The use of distributional word embeddings to compare meanings is based on the ability of these models to capture latent information present in texts, as shown by the works of \cite{Bolukbasi2016debiasing} and  \cite{caliskan2017semantics}, where researchers show that distributed representations contain human-like biases.

On the other hand, a series of different works generalize the study of biases to the analysis of semantic differences in collections over various dimensions such as geographically situated language~\cite{galliani2019}, diachronic collections \cite{Kulkarni2015StatisticallyChange,Hamilton2016DiachronicChange,Yao2017DiscoveryLearning,garg2018word,Tripodi2019antisemitic,hamilton2016cultural}, and collections of different sources~\cite{hamilton2016inducing,galliani2019}.
The nature of the studied semantic differences was also expanded, encompassing gender~\cite{Zhao2018LearningGW,garg2018word}, racial and ethnic stereotypes~\cite{garg2018word,Tripodi2019antisemitic,galliani2019}, and sentiment analysis \cite{hamilton2016inducing}.

In these contexts, two groups of approaches can be identified; those who treat meaning shift and semantic differences --- that is the main theme of this research work --- and those that use wordsets to identify biases in language.

\paragraph{Meaning Shift}
For aligned representations, it is possible to investigate the semantic change of a word by simply comparing the representations, which will be examined further. The context of diachronic collections is one of the most common applications for aligned representation models and the evolution of word usage is often the object of interest in the literature~\cite{Kulkarni2015StatisticallyChange,Hamilton2016DiachronicChange}. While most methods use simple vector similarity, others use first neighborhood search \cite{Yao2017DiscoveryLearning}. A hybrid approach is the one of~\cite{galliani2019} where a wordset-based approach is used on aligned representation, allowing for comparative semantic analysis on both diachronic and geographical dimensions.

\paragraph{Wordset Based}
The original methodology introduced by~\cite{caliskan2017semantics} to measure bias  investigates the association for two sets of opposite target words (e.g., \textit{scientific} and \textit{artistic} professions) against two sets of polarising attribute words (e.g., \textit{male}-like and \textit{female}-like words). 
Other works such as~\cite{garg2018word} generalize the previous methodology and replace wordsets with their respective average representation and the second target set with its complement to background, meaning all words that do not exhibit some particular connotation (e.g., \textit{immigrant}-like terms and everything else), enabling for single-target analyses. Another approach is the one used in~\cite{hamilton2016inducing}, where a label propagation algorithm is used to induce a sentiment score from a seed lexicon (wordset) to the graph of neighbors. All these techniques do not require an alignment of the representations since they use a computed score for indirect comparison instead.


\subsection{Temporal Word Embeddings}\label{sec:termoral:word:embeddings}

The alignment problem of distributional representations has been deeply studied in the field of temporal word embeddings, in which researchers want to create vector representations of words in different periods of time to analyze semantic change~\cite{Hamilton2016DiachronicChange,Kulkarni2015StatisticallyChange,rudolph2016exponential} (see \cite{boleda2019distributional} for a brief overview of the semantic change topic).

A \ac{twem} is a model that learns \textit{temporal word embeddings}, i.e., vectors that represent the meaning of words during a specific temporal interval. For example, a \ac{twem} is expected to associate different vectors with the word \textit{gay} at different times: its vector in the representation of the year $1900$ is expected to be more similar to the vector of terms like \textit{joyful} than its vector in $2005$. By building a sequence of temporal embeddings of a word over consecutive time intervals, one can track the semantic shift in meaning that occurs in the word usage.

Most of the proposed \ac{twem}s align multiple vector spaces by enforcing word embeddings in different time periods to be similar \cite{Kulkarni2015StatisticallyChange,Rudolph2017DynamicEvolution}.
The underlying assumption is that the majority of the words do not change their meaning over time.  
This approach is reasonable but may be misleading for some words, as it can excessively smoothen differences between meanings that have shifted along time.
A remarkable limitation of current \ac{twem}s is related to the assumptions they make on the size of the corpus needed for training: while some methods like~\cite{Szymanski2017TemporalEmbeddings,Hamilton2016DiachronicChange} require a huge amount of training data, which may be difficult to acquire in several application domains. Other methods like~\cite{Yao2017DiscoveryLearning,Rudolph2017DynamicEvolution} may not scale well when trained on big datasets.

Different researchers have investigated the use of word embeddings to analyze the semantic changes of words over time~\cite{Hamilton2016DiachronicChange,Kulkarni2015StatisticallyChange}.
We identify two main groups of approaches that are based on the strategy applied to align temporal word embeddings associated with different time periods: one is referred to as \textit{pairwise alignment} in which pairs of vector spaces are aligned, while the second is referred to as \textit{joint alignment} in which global constraints on the optimization process are used to generate vector spaces that are aligned after training.

\paragraph{Pairwise Alignment}
Pairwise Alignment-based approaches align pairs of vector spaces to a unique coordinate system: \cite{Kim2014TemporalModels} and \cite{DelTredici2016TracingSpaces} align consecutive temporal vectors through neural network initialization; other authors apply various linear transformations after training that minimize the distance between the pairs of vectors associated with each word in two vector spaces \cite{Kulkarni2015StatisticallyChange,Hamilton2016DiachronicChange,Szymanski2017TemporalEmbeddings,Zhang2016TheTime}. Essentially what can be learned is a matrix $\mathbf{M}_{s^1,s^2}$ that maps words from the vector space $s_1$ to the vector space $s_2$.

\paragraph{Joint alignment}

Joint alignment-based approaches train all the temporal vectors concurrently, constricting them to a unique coordinate system:  \cite{Bamman2014DistributedLanguage} extend Skip-gram Word2vec tying all the temporal embeddings of a word to a common global vector (they originally apply this method to detect geographical language variations); other models impose constraints on consecutive vectors in the Positive Point-wise Mutual Information (PPMI) matrix factorization process \cite{Yao2017DiscoveryLearning} or when training probabilistic models to enforce the ``smoothness'' of the vectors' trajectory along time~\cite{Bamler2017DynamicEmbeddings,Rudolph2017DynamicEvolution}. This strategy leads to better embeddings when smaller corpora are used for training but it is less efficient then pairwise alignment.

\paragraph{Comparison}
Despite the differences between the pairwise and the joint alignment strategies, both strategies try to enforce the vector similarities among different temporal embeddings associated with the same word.
While this alignment principle is well-motivated from a theoretical and practical point of view, enforcing the vector similarity of one word across time may lead to excessively smoothen the differences between its representations in different time periods.
Finding a good balance between \textit{dynamism} and \textit{staticness} is an important feature of a \ac{twem}. Finally, note that very few models proposed in the literature
do not currently require explicit pairwise or joint alignment of the vectors, and these models all rely on co-occurrence matrix or high-dimensional vectors \cite{Gulordava2011ACorpus,Basile2016DiachronicNgram}. Consider that these strategies assume temporal continuity. Note that wordset approaches are supervised and generally require the definition of specific lexicons, thus making it impossible to have an implicit comparative framework.

\subsection{Cross-lingual Embeddings}
In this comparison with the literature, it is also important to cite multilingual approaches that are meant to find mappings between words (or sentences) of different languages. There are many different works that try to generate aligned representations by first defining anchor points that should not move in the space: a set of reference coordinates to align everything, this is common in the multilingual alignment community, in which the objective is to generate aligned representations of different languages. Multilingual lexicons are used to stabilize the position of some words in the vector space~\cite{faruqui2014improving,xing2015normalized,Smith2017ODictionary}: defining a set of anchors or a mapping dictionary requires domain knowledge and might be challenging in some contexts since it requires apriori domain knowledge. Thus, Facebook proposed an approach to align multilingual corpora without lexicon, this approach, named MUSE~\cite{conneau2017word},  leverages on adversarial training, to align different multilingual corpora. We refer the reader to~\cite{ruder2019survey} for an in depth analysis of multi-lingual word embeddings.

\section{Compass-aligned Distributional Embeddings}\label{sec:compass}

In this Section, we describe a framework for the comparative analysis of meanings of words used in different corpora. The framework is based on word-level embeddings, which means that it addresses differences in core meanings of words, as also proposed in previous work~\cite{hamilton2016cultural}. 
Table~\ref{tab:example:comparison} shows some examples of possible comparisons that we would like to make between different corpora in two different domains. 
One domain considers the study of linguistic differences by comparing words' meaning in American and British English based on two representative news corpora, e.g., one from The New York Times and one from The Guardian (see Section~\ref{sec:experiments:generalization} for details about these corpora). 
Another domain considers the study of temporal differences by comparing words' meaning in corpora from different time periods, e.g., New York Times articles written in 1987 vs. 2007 (see Section~\ref{sec:experiments:temporal:analogies} for details about these corpora). 
The comparisons reported in the examples are based on semantic correspondences, that is cross-corpora linguistic equivalences to identify which words have the same usage in two distinct corpora. 
These correspondences can be also viewed as cross-corpora analogies: e.g., "flat" is the English word whose meaning is closest to the meaning of the American word "apartment"; "clinton" is the word whose meaning in 1997 is closest to the meaning of "reagan" in 1987. 
The objective of the framework is to support comparison in different domains, both in the case where an obvious order can be imputed to the considered corpora, e.g., based on the contiguity of the time periods associated with each corpus in a temporal domain~\cite{Hamilton2016DiachronicChange,Yao2017DiscoveryLearning}, and in the case where such an order can or may not be imputed, e.g., in domains identified by topics such as games, business, politics, and so on~\cite{Bamman2014DistributedLanguage}.

\begin{table}[]
    \centering
    \small
    \begin{tabular}{cccc} \toprule
         Source Corpus (D1) & Input Word (D1) & Target Corpus (D2) & Output word (D2)  \\ \midrule
         US & apartment & UK & \textbf{flat} \\
         US & labeled & UK & \textbf{labelled} \\
         US & gasoline & UK & \textbf{petrol} \\
         1987 & reagan & 1997 & \textbf{clinton} \\
         1987 & walkman & 2007 & \textbf{ipod} \\ \bottomrule
    \end{tabular}
    \caption{Some example of comparisons we would like a comparative model to be able to find find. These examples can also be viewed as analogies between different corpora~\cite{Szymanski2017TemporalEmbeddings}.}
    \label{tab:example:comparison}
\end{table}{}

To achieve this domain-agnostic objective, the Compass-aligned Distributional Embeddings (CADE) produce aligned distributional representations of words from a collection of corpora, capturing the semantic differences in word usage as differences in representations which can then be explored and quantified thanks to the alignment.
In the next sections, after establishing some preliminaries concerning the word2vec model, first, the comparison framework is introduced together with its notation and definitions and later the alignment method used in \ac{atmodel} is presented.

\subsection{Preliminaries: Target and Context Matrices in word2vec}

The word2vec algorithm~\cite{Mikolov2013DistributedCompositionality} uses a feed-forward neural network architecture; in the original work two models were presented, Continous Bag of Word (CBOW) and Skip-gram. 
The difference between the two is the task on which they are trained: the first, CBOW, aims to predict a target word given its contexts --- i.e., it's neighbors --- while the second,  Skip-gram, tries to predict, given a target word, its context.

\begin{figure*}[h]
\centering
\includegraphics[width=0.5\textwidth]{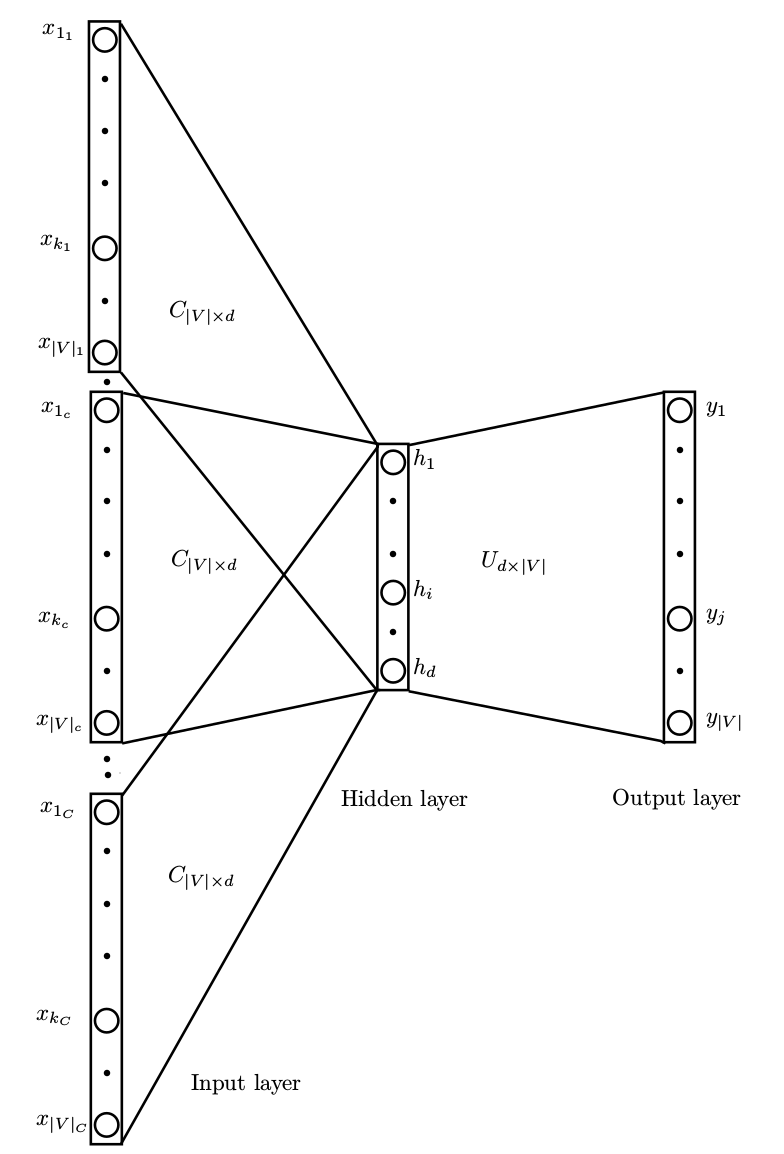}
\caption{A schematic representation of the CBOW model.}
\label{fig:cbow} 
\end{figure*}

Since word2vec architecture is a two-layer feed-forward neural network, it comes with two matrices: one that projects inputs to a hidden space and the second one that projects to the output space. 
For example, CBOW comes with a context matrix $\textbf{C}$ and a a target matrix $\textbf{U}$. 
After training, the word embeddings are found in the context matrix $\textbf{C}$.
See Figure~\ref{fig:cbow} for a schematic representation of the CBOW model, we show that there are two matrices ($\textbf{C}$ and $\textbf{U}$), where $\textbf{C}$ project the input in the hidden layer --- the embeddings --- and $\textbf{U}$ project the from the hidden layer to the output layer.

Once the vector representations are generated, it is possible to compute the similarity between different words using measures such as cosine similarity or the Euclidean distance. 
In this work we will follow the literature consensus by computing the similarity between words using the cosine similarity between their embedding vectors but this is by no means the only possible choice.

\subsection{Comparison Framework for Distributional Models}

In our comparative framework we have access to a \textbf{collection} $D$ that consists of various corpora $D^1, \ldots, D^n$. 
When considered as part of a collection, we refer to each corpus $D^i$ as to the \textbf{slice} $i$ of the collection. 
In general, we regard $i$ as a plain index, or, name, for a slice, with the order $1,...,n$ not reflecting a specific order relevant for comparison. 
In some domains, it could be convenient to index the slices to reflect an order implicit in the discriminant used to split the collection, e.g., in the temporal domain, with a set of corpora $D^{1985}, D^{1986}\ldots, D^{2010}$. 
Given a collection $D$, with $V$ we refer to its \textit{global vocabulary}, that is, the set of tokens that occur in $D$, and with $w_x$ we refer the token $x$ of the vocabulary, e.g., $w_{apartment}$. For simplicity we may also refer to the token $w_x$ simply as $\text{x}$, e.g., $w_{apartment}=\text{``apartment''}$.  
We refer to \textit{tokens} rather than \textit{words} to highlight that the framework is general enough to be used with any adaptation of word2vec, including the ones that generate embeddings also for word phrases containing more words, e.g., where a token such as ``ronald reagan'' would be part of $V$, or the ones that generate embeddings for entity identifiers, e.g., where \url{http://dbpedia.org/resource/Ronald_Reagan} would be part of $V$. 
However, in the following we will focus on word embeddings and use the terms ``word'' and ``token'' interchangeably. 
Different corpora may contain proper subsets of the global vocabulary, with some token being present in a limited number of corpora. With $V^i$ we refer to the vocabulary restricted to the slice $D^i$, i.e, the subset of $V$ that contains only tokens that occur also in $D^i$. 
For example, consider a collection of news articles $NA = \{D^{NYT},D^{GUA}\}$, where $D^{NYT}$ and $D^{GUA}$ represent the slices of articles extracted from The New York Times and The Guardian respectively; $V^{NYT}$ and $V^{GUA}$ contains only the tokens that occur respectively in $D^{NYT}$ and $D^{GUA}$, with $V^{NA} = V^{NYT} \cup V^{GUA}$.  

Each corpus $D^i$ is associated with a set of \textbf{slice-specific embeddings} $\textbf{C}^i \subseteq \mathbb{R}^h$ (or, also, corpus-specific embeddings), each one consisting of the vectors associated with the tokens in $V^i$, for a dimension $h$ that is shared across the collection. 
In the previous example we consider the two sets of embeddings $\textbf{C}^{NYT}$ and $\textbf{C}^{GUA}$ associated respectively to the slices containing articles from the New York Times and the Guardian. 
We use the notation $\textbf{c}_{x}^i$ to refer to the \textbf{slice-specific vector} associated to the $x$-token of $V$ in the $\textbf{C}^i$ embeddings. 
For example, $\textbf{c}_{apartment}^{NYT}$ refers to the NYT-specific vector of the token ``apartment''. 
We refer to this vector also as to the $i$-th slice-specific vector of the token $w_x$, and we refer to $\textbf{C}^i$ as to $i$-th slice-specific embeddings, that is, the set of the slice-specific vectors associated with the tokens in $V^i$. 
When we do not need to refer to a specific word, we will also use the simplified notation $\textbf{c}^{i}$ to refer to the vector of a generic word $w$ in  $\textbf{C}^i$. 
Finally, we define \textbf{collection embeddings}, denoted as $\textbf{C}$, as the set union of all the slice-specific embeddings $C^i$. For example, the embeddings for the news articles' collection $NA$ is referred to as $\textbf{NA}=\textbf{C}^{NYT} \cup \textbf{C}^{GUA}$. 

Observe that while we can define a bijection between $V^i$ to $\textbf{C}^i$, i.e., each token that occurs in a slice $i$ is associated with a slice-specific vector, the mapping between $V$ and a slice-specific embedding $\textbf{C}^i$ is partial because a word $w \in V$ may not occur in $V^i$ and thus not have a slice-specific vector in $\textbf{C}^i$. Observe also that indexes are defined in $V$ and shared across slices, i.e., $w_x$ refers to the same token across the slices, even though $w_k$ may not be part of some slice-restricted vocabulary.

It is worth remarking that all slice-specific embeddings in the collection reside in the same vector space $\mathbb{R}^h$: this enables the comparison of their corpus-specific vectors using operations (such as cosine similarity) defined on vector space elements, thus bypassing the distinction between slices. For example $cosine(\textbf{c}_{apartment}^{NYT},\textbf{c}_{apartment}^{GUA})$ evaluates the cosine similarity between the the NYT-specific and GUA-specific vectors associated with the word ``apartment''.
While these comparisons would always be possible between vectors having the same dimension, they are only meaningful when the slice-specific vectors are well-aligned. 

Comparisons across aligned embeddings can be implemented using well-known similarity and distance measures over vector spaces as building blocks, and, in particular, cosine similarity and Euclidean distance. However, we introduce a few discrete comparison functions that support intuitive cross-corpora semantic difference analysis based on a chosen similarity measure.       

We start by introducing a \textbf{correspondence function} that maps each token in a source slice to a (possibly different) token in a target slice. The function is defined by evaluating the similarity between corpus-specific vectors, i.e., it maps an input token to an output token based on the vectors that represent the tokens in the source and target slices. Intuitively, it finds the word whose usage in a target corpus is evaluated to be most similar to the usage of the input word in the source corpus (see Table~\ref{tab:example:comparison} for examples).

\begin{idefinition}[Cross-corpora Word Correspondence]
Given two slices $D^{i}$ and $D^{j}$ with vocabularies $V^i$ and $V^j$ and a similarity measure $\sigma$, we define a correspondence function $\phi_{D^{i} \rightarrow D^{j}}$ as a function that for every token $w_x \in V^i$ associates a token $w_y \in V^j$ if and only if $\textbf{c}_{y}^i$ is the most $\sigma$-similar vector to the vector $\textbf{c}_{x}^i$.
\end{idefinition}

\begin{figure*}[h]
\centering
\includegraphics[width=0.75\textwidth]{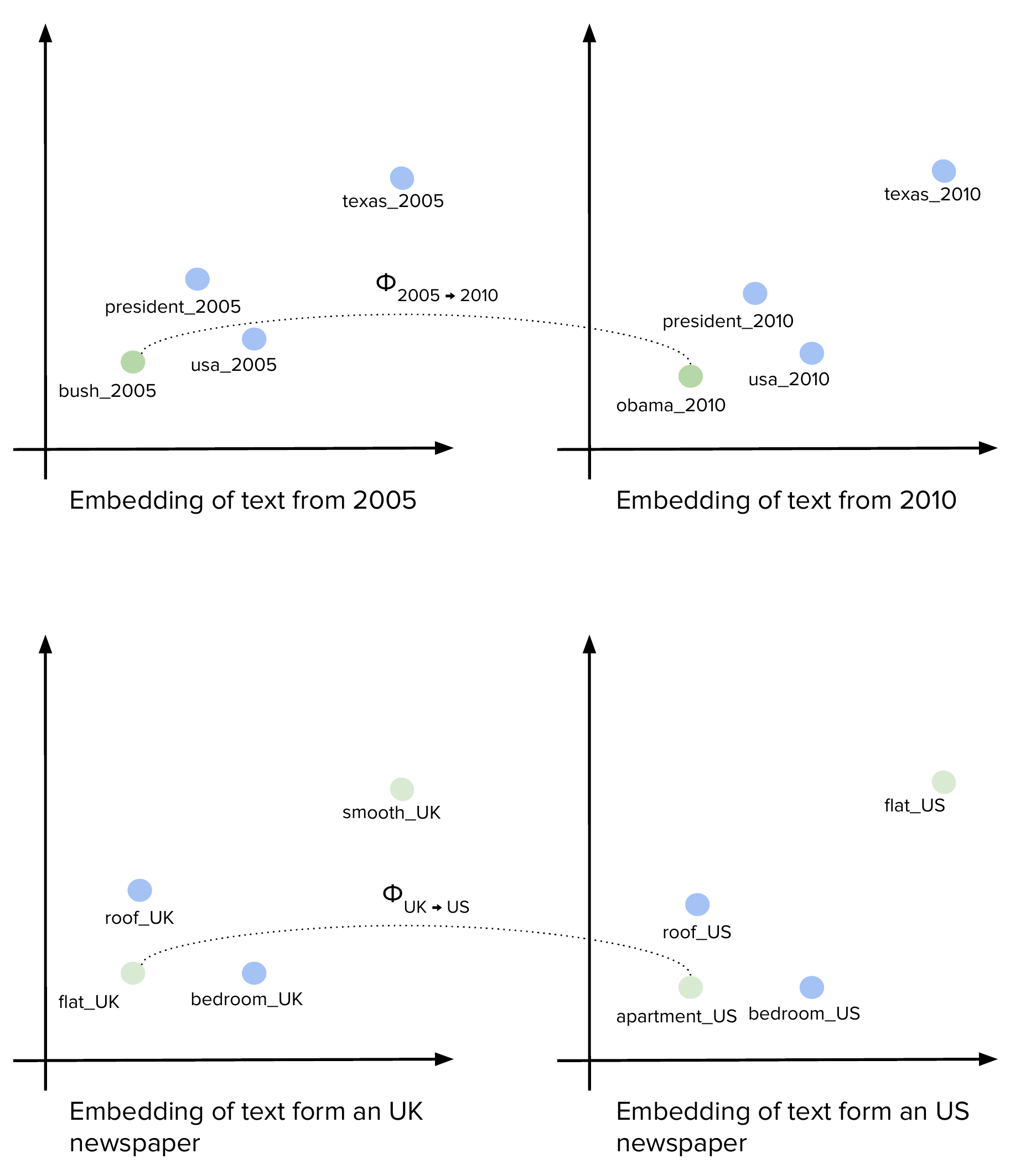}
\caption{Examples of cross-corpora correspondence functions and significant changes in the vectors' position in a 2D projection; a simplified notation is used to indicate corpus specific vectors' projections, e.g., $\text{obama\_2010}$ refers to the projection of $\textbf{c}^{2010}_{obama}$ . }
\label{fig:cade:langauges:example} 
\end{figure*}

In the rest of this work we will use cosine similarity as similarity measure $\sigma$ because it is not affected by the magnitude of the vectors, which in word embeddings is sensitive to word frequency~\cite{schakel2015measuring,wendlandt2018factors}. More examples of correspondence functions across aligned corpora in two domains, the temporal and the language localization domain, are represented in Figure~\ref{fig:cade:langauges:example}, where the simplified notation $\text{x\_i}$ is used to label the 2D projection of a vector $\textbf{c}^i_{x}$ and just a portion of the 2D projection of the embedding is shown. Note that the positions of the vectors have changed slightly for some words (highlighted in light blue), e.g.,``president" in the temporal domain and ``roof" in the language localization domain, and sharply for other words (highlighted in light green), e.g., ``flat" in the language localization domain, some of which would appear outside the space shown in the figure.

This correspondence function can be used straight away to model \textbf{cross-corpora analogies}, a generalization of temporal analogies often used to test the performance of temporal word embeddings. A cross-corpora analogy can be expressed in a propositional form as $w_x$ : $D^i$ :: $w_y$ : $D^j$, which could be read as ``$w_x$ is in (the context of) $D^i$, what $w_y$ is in (the context of) $D^j$", with the corpora as representative of aggregated contexts. Clearly, such analogy translates into the correspondence $\phi_{D^{i} \rightarrow D^{j}}(w_x)=w_y$. Examples of cross-corpora analogies can be easily derived from the examples of correspondence represented in Figure~\ref{fig:cade:langauges:example}, e.g., ``apartment'' is to American-English what ``flat'' is to British-English.      

We remark that there are two possible outcomes from the application of a correspondence function, where $\phi_{D^{i} \rightarrow D^{j}}(w_x)=w_y$: for $x=y$, the meaning of the token $w_x$ is deemed to be stable across the corpora $D^{i}$ and $D^{i}$, while for $x=y$, the meaning of the token $w_x$ is deemed to have changed. The correspondence function can be viewed as a discrete measure of change, but it can be quite sensitive to small changes. For this reason, it is often convenient to look not only to semantic correspondences but to a larger neighborhood of words that have a similar meaning in the target corpus. For this reason, it is useful to generalize the correspondence functions into a family of functions that retrieve top-k nearest neighbors across corpora as follows.

\begin{idefinition}[Cross-corpora Top-k Nearest-Neighbours]
Given two slices $D^{i}$ and $D^{j}$ with slice-restricted vocabularies $V^i$ and $V^j$, we define the correspondence function  $\phi^{k}_{D^{i} \rightarrow D^{j}}$ as a function that map every token $w_x \in V^i$ to the set of $k$ tokens $w_y \in V^j$ whose slice-specific vectors $\textbf{c}_y^j$ are the vectors in $\textbf{C}^j$ that are most similar to the vector $\textbf{c}_{x}^i$.
\end{idefinition}

Obviously, a correspondence function, denoted by $\phi_{D^{i} \rightarrow D^{j}}$, is equivalent to the cross-corpora top-1 nearest neighbour function $\phi^{1}_{D^{i} \rightarrow D^{j}}$. We can summarize the above definitions that define a framework to support cross-corpora semantic comparison across aligned word embeddings with the definition of the comparative distributional framework. 

\begin{idefinition}[Comparative Distributional Framework]
A comparative distributional framework is a quadruple $\mathcal{F} = (D,V^{*},\textbf{C},\Phi)$, where: $D = \{D^1,\ldots,D^n\}$ is a collection of slices; $V^{*}=\{V, V^1, \ldots, V^n\}$ is the set of vocabularies that include the global vocabulary $V$ and all the slice-restricted vocabularies $V^i$, such that each $V^i\subseteq V$ is limited to word occurrences in the $i$-th slice and $V = \bigcup_{i=1}^n V^i$; $\textbf{C} = \bigcup_{i=1}^n \textbf{C}^{i}$ is the union of a set of slice-specific embeddings $\textbf{C}^i$, each one generated from the slice $D^i$ and aligned to all the slice-specific embeddings $\textbf{C}^j$ with $i\neq j$; $\Phi$ is a family of cross-corpora top-k nearest-neighbours functions $\phi^{k}_{D^i \rightarrow D^j}$ defined for all $i$ and $j$, i.e., between every pair of slices $D^i$ and $D^j$ in the collection.
\end{idefinition}

\subsection{Compass-aligned Distributional Embeddings}

With CADE we refer to the compass-based method used to return a set of pairwise aligned slice-specific embeddings, which can support a comparative distributional framework. 
This method takes a collection of slices as input and returns the slice-specific embeddings, each of which is aligned with all the other slice-specific embeddings. As a result, all pairs of embeddings are aligned and are embedded in the same vector-space, thus supporting the pair-wise comparison operations defined in the comparison framework.

Our approach is inspired by an assumption made in previous work by \citeauthor{Kulkarni2015StatisticallyChange}: the majority of words do not change their meaning over time. 
While some words assume different meanings over time (e.g., ``amazon'', ``apple'', ``gay'') most of the words tend to have a stable meaning. 
Nevertheless, we believe that this assumption is also true for other aspects: two sources that use the same language might use the same word differently (i.e., think of the differences between British and American English), but most of the words used for communication have a strong shared meaning.

From this assumption, we derive a second one: we assume that a shifted word, i.e., a word whose meaning has changed, appears in the contexts of words whose meaning changes only slightly. However, our assumption is particularly true for shifted words: for example, the word \textit{clinton} appears during some time periods in the contexts of words that are related to his position as president of the USA (e.g., \textit{president}, \textit{administration}); conversely, the meanings of these words have not changed.  The same assumption can be applied to the word ``petrol'', used in British English (i.e., the American English equivalent is ``gas''). This word will appear in contexts related to ``cars'' and ``economy'', words that have a stabler meaning.

The above assumptions allow us to heuristically consider the target embeddings as static, i.e., to freeze them during training, while allowing the context embeddings to change based on co-occurrence frequencies that are specific to a given slice. Thus, our training method returns the context embeddings as word embeddings. 
 
Finally, we observe that our compass method can be applied also in the opposite way, i.e., by freezing the context embeddings and moving the target embeddings, which are eventually returned as the word embeddings. However, a thorough comparison between these two specular compass-based training strategies is out of the scope of this paper.

\begin{figure*}[h]
\centering
\includegraphics[width=0.95\textwidth]{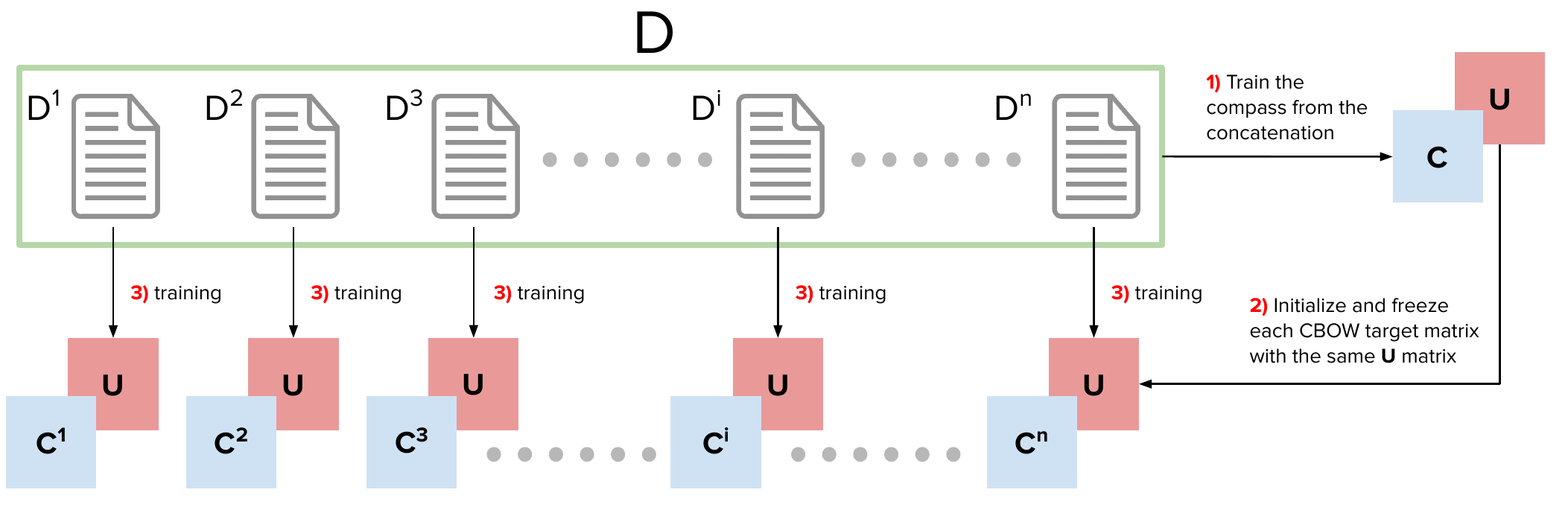}
\caption{The \ac{atmodel} model.}
\label{fig:cade} 
\end{figure*}

As we said, \ac{atmodel} can be implemented on top of the two Word2vec models,  Skip-gram and CBOW. Here we present the details of our model using CBOW as the base Word2vec model, since we empirically found that it produces models that show better performance than Skip-gram with small datasets.  The training process of \ac{atmodel} is divided in three phases, which are schematically depicted in Figure \ref{fig:cade}. 

(1) First, we construct two \textit{compass} matrices $\mathbf{C}$ and $\mathbf{U}$ by applying the original CBOW model on the whole corpus $D$; $\mathbf{C}$ and $\mathbf{U}$ represents the set of \textit{compass context embeddings} and \textit{compass target embeddings}, respectively. We discard the $\mathbf{C}$ matrix.
(2) Second, for each specific slice $D^{i}$, we construct the context embedding matrix $\mathbf{C}^{i}$ as follows. We initialize the output weight matrix of the neural network with the previously trained compass target embeddings from the matrix $\mathbf{U}$, the $\mathbf{C}^{i}$ matrix is initialized as in ~\citeauthor{Mikolov2013DistributedCompositionality}. (3) We run the CBOW algorithm on the specific slice $D^{i}$ and during this training process, the target embeddings of the output matrix $\mathbf{U}$ are not modified (i.e., we \textit{freeze} the layer), while we update the context embeddings in the input matrix $\mathbf{C}^{i}$. After applying this process on all the slices $D^{i}$, each input matrix $\mathbf{C}^{i}$ will represent our word embeddings for the slice $i$. Here below we further explain the key phase in our model, that is, the update of the input matrix for each slice, and the interpretation of the update function in our model. 

Given a slice $D^{i}$, the second phase of the training process can be formalized for a single training sample $\langle w_{x},\gamma(w_{x}) \rangle \in D^{i}$ as the following optimization problem:
\begin{equation}
\begin{split}
\max_{\mathbf{C}^i} \log P(w_{x}|\gamma(w_{x})) =  \sigma(\textbf{u}_{x} \cdot \textbf{c}_{\gamma(w_{x})}^{\;i}) 
\end{split}\label{loss}
\end{equation}
where $\gamma(w_x)=\langle w_{1},\cdots,w_{j} \rangle$ represents the words in the context of $w_x$ which appear in $D^i$, $\textbf{u}_x \in \mathbf{U}$ is the compass target embedding of the word $w_x$, and 
\begin{equation}
\begin{split}
\textbf{c}_{\gamma(w_{x})}^{\;i} = \frac{1}{|\gamma(w_x)|} (\textbf{c}_{1}^{\;i} + \cdots + \textbf{c}_{j}^{\;i})^T\\
\end{split}\label{c_cw}
\end{equation}
is the mean of the context embeddings of the contextual words $w_{1 \ldots j}$. The softmax function $\sigma$ is calculated using Negative Sampling \cite{Mikolov2013DistributedCompositionality}. Please note that $\mathbf{C}^i$ is the only weight matrix to be optimized in this phase ($\mathbf{U}$ is constant), which is the main difference from the classic CBOW. The training process maximizes the probability that given the context of a word $w_x$ in a particular slice $i$, we can predict that word using its target matrix $\mathbf{U}$. Intuitively, it moves the slice-specific context embedding $\textbf{c}_{j}^{\;i}$ closer to the compass target embeddings $\textbf{u}_{x}$ of the words that usually have the word $w_{j}$ in their contexts in slice $i$. 
The resulting context embeddings can be used as word embeddings: they will be already aligned, thanks to the shared compass target embeddings used as a compass during the independent training.

For example, we update the slice-specific representation of the token "obama" when this token appears in the context of a target token like "president" or "barak".

Observe that our model comes in two flavors: before step (3) of the process, the matrix $C^i$ can be initialized with the vectors of $C$, the compass matrix. At the same time, we can randomly initialize the weights of matrix $C^i$ and fine-tune them accordingly to the text of the slice. The first setting is much more conservative since the vectors start with an already trained embedding it is more difficult to move them. In the second case, the training allows us to build vectors that are entirely based on the slice considered, but in case of slices with few textual information (i.e., non-representative text) this might skew the results. The setting to use depends on the context on which these embeddings should be used. We will make use of the latter approach in most experiments.

We observe that differently from those approaches that enforce similarity between consecutive word embeddings (for example in the temporal domain  \citeauthor{Rudolph2017DynamicEvolution}), \ac{atmodel} does not apply any slice-specific assumption.

The proposed method can be viewed as a method that implements the main intuition of \cite{Gulordava2011ACorpus} using neural networks, and as a simplification of the models of \cite{Rudolph2017DynamicEvolution,Bamman2014DistributedLanguage}. Despite this simplification, experiments show that \ac{atmodel} outperforms or equals more sophisticated versions on different experimental evaluations. 
Our model has the same complexity of CBOW over the concatenation of the slices that are in $D$, plus the task of computing $n$ CBOW models over all the slices. Note that this last training can be run in parallel since the training of each slice is independent of the others.

\subsection{Open Sourcing CADE}
We provide CADE as an open-source platform to align distributional embeddings\footnote{\url{http://github.com/vinid/cade}}. Our tool is based on the well known Gensim\footnote{\url{https://radimrehurek.com/gensim}} library and thus it automatically inherits all the properties and the methods. We also provide documentation on how to use this tool and experiment with it. This tool is easy to use also for people outside the computer science community: little knowledge about programming and word embeddings is necessary to deal with our tool. We think this is useful also for those communities outside computer science that in recent years have started to strongly rely on word embeddings like psychology or psycho-linguistics.
\section{Experimental Evaluation: Objectives and Overview}\label{experiments}

Our deeper experimental evaluation is based on temporal word embeddings; we mainly focus on this category for two reasons: (i) temporal word embeddings are a field that is getting much interest lately~\cite{Yao2017DiscoveryLearning,Rudolph2017DynamicEvolution,Kulkarni2015StatisticallyChange,Hamilton2016DiachronicChange} and thus, the experimental evaluation and datasets are now standardized; moreover (ii), we believe that temporal data is a good prototype to model language differences: chronologically close corpora tend to be more similar than chronologically distant corpora and thus, dealing with temporal data offers the possibility of evaluating more general characteristics of language change. Section~\ref{sec:experiments:temporal:analogies} contains these experiments.

To summarize, the main motivations that drove this experiment were:
\begin{itemize}
    \item there do exist well defined experimental settings and the possibility of comparison with respect to different methods;
    \item temporal evolution is a challenging problem, where trajectories across several independent slices may be challenging to find for a method like ours that does not directly consider interdependencies between the corpora;
    \item this domain is one of the most studied and relevant in which meaning shift is analyzed (evolution of language).
\end{itemize}{}

After this first experimental evaluation, we show that our model can be generalized on non-temporal data. These experiments are found in Section~\ref{sec:experiments:generalization} where we show that we can compare American and British English. We show different examples with the comparison of corpora with different topics, showing at the same time for which tasks \ac{atmodel} can be used.

Eventually in Section~\ref{sec:experiments:robustness}, we evaluate the robustness of the model. Robustness of the model is evaluated by looking at two main issues: how much is our approach sensitive to semantic change and how the vocabulary overlap across corpora affects the performance of \ac{atmodel}.  
Two aspects of \ac{atmodel}'s sensitivity to change are evaluated. The first aspect we evaluate is the capability of \ac{atmodel} to detect change, which is aimed at showing that compared to other approaches \ac{atmodel} finds a good trade-off between overestimating and underestimating change; this experiment thus provides some explanation of the results discussed in Section~\ref{sec:experiments:temporal:analogies} and is based on the temporal analogies used in the previous experiments. The second aspect we evaluate is the robustness of \ac{atmodel} against increasing amount of change, which is simulated in controlled settings using a sample of the Guardian corpus also used in Section\ref{sec:experiments:generalization}. 
The experiment about impact of vocabulary overlap is aimed at providing better insights on a factor that is important to ensure the quality of the \ac{atmodel} alignment: the overlap of a significant portion of the vocabulary across slices; in particular, we will discuss how the quality of the alignment is affected by controlled modification of the degree of overlap.

\section{Performance: Experiments on Temporal Word Embeddings}\label{sec:experiments:temporal:analogies}

\paragraph{Overview.}

We compare \ac{atmodel} with static and with state-of-the-art models that have shown better performance according to the literature of Temporal Word Embedding Models (TWEM). We have used two methodologies proposed to evaluate temporal embeddings so far: temporal analogical reasoning~\cite{Yao2017DiscoveryLearning} and held-out tests~\cite{Rudolph2017DynamicEvolution}.
\subsection{Experiments with Temporal Analogies}

To evaluate \ac{atmodel} we focus on two different datasets. Each dataset consists of a collection of corpora and a set of temporal analogies. The difference between these two datasets lies in the size of the collection and the amount of available training data.

\subsubsection{Datasets and Methodology} The \textit{small dataset} \cite{Yao2017DiscoveryLearning} is freely available online\footnote{https://sites.google.com/site/zijunyaorutgers/publications}.
We will refer to this dataset as \ac{datas}. The \textit{big dataset} is the New York Times Annotated Corpus\footnote{https://catalog.ldc.upenn.edu/ldc2008t19}  \cite{Sandhaus2008TheCorpus} employed by \citeauthor{Szymanski2017TemporalEmbeddings,Zhang2016TheTime} to evaluate their models. We will refer to this dataset as \ac{datab}. Both datasets are divided into slices, each containing one year of data.  \ac{tests} introduced by \citeauthor{Yao2017DiscoveryLearning} and \ac{testb} introduced by \citeauthor{Szymanski2017TemporalEmbeddings}.
They are both composed of temporal word analogies based on publicly recorded knowledge, partitioned in categories (e.g., \textit{President of the USA, Super Bowl Champions}).  Numeric information about datasets and test sets are summarized in Table \ref{test:tabledata}.

\begin{table}[]
\centering
\small
\begin{tabular}{ccccccc}
\toprule
Data  & Words &  Span & Slices   \\ \midrule
\ac{datas}  &
$50$M &  $1990$-$2016$ & $27$ \\ 
\ac{datab}  & $668$M
 &  $1987$-$2007$ & $21$  \\ 
 \acs{dataml}  & $6.5$M &  $2007$-$2015$ & $9$
 \\ \toprule
Test &  Analogies &  Span & Categories \\ \midrule
\ac{tests}  &
$11,028$ & $1990$-$2016$ & $25$ \\ 
\ac{testb}  &
 $4,200$ &  $1987$-$2007$ & $10$  \\ \bottomrule
\end{tabular}%
\caption{Details of \ac{datas}, \ac{datab}, \acs{dataml}, \ac{tests} and \ac{testb}.}
\label{test:tabledata}
\end{table}

To test the models trained on \ac{datas} we used the \ac{tests}, while to test the models trained on \ac{datab} we used the \ac{testb}. This allows us to replicate the settings of the work of \citeauthor{Yao2017DiscoveryLearning} and \citeauthor{Szymanski2017TemporalEmbeddings} respectively.

We extend the analysis on the analogies by studying the results under a deeper point of view: given an analogy $w_1 : t_1 = x : t_2$, we define \textit{time depth} $\delta_t$ as the distance between the temporal intervals involved in the analogy: $\delta_t=|t_1-t_2|$.
Analogies can be divided in two subsets: set $Static$ consists of \textit{static analogies}, which involve a pair of the same words (\textit{obama : 2009 = obama : 2010}), and the set $Dynamic$ consists of \textit{dynamic analogies}, that are not static. We refer to the complete set of analogies as $All$. Given a model and a set of temporal analogies, the evaluation of the given answer is done with the use of two standard metrics, the Mean Reciprocal Rank (MRR) and Mean Precision at K (MP@K).

\subsubsection{Baselines}
We tested different models to compare the results of \ac{atmodel} with the ones provided by the literature: two models that apply pairwise alignment, two models that apply joint alignment and a baseline static model. Where not differently stated, we implemented them with CBOW and Negative Sampling extending the \textit{gensim} library. We compare \ac{atmodel} with the following models:
\begin{itemize}
\item \textit{LinearTrans-Word2vec} (\textit{TW2V}) \cite{Szymanski2017TemporalEmbeddings}.
\item \textit{OrthoTrans-Word2vec} (\textit{OW2V}) \cite{Hamilton2016DiachronicChange}.
\item \textit{Dynamic-Word2vec} (\textit{DW2V}) \cite{Yao2017DiscoveryLearning}. There are some issues on the coding repository web page that prevent us from completely replicating the experiment. However, the authors provided the dataset and the test set of their evaluation settings (the same employed in our experiments) and published their results using our same metrics. Thus, we also included DW2V into the experimental evaluation.
\item \textit{Geo-Word2vec} (\textit{GW2V})
\cite{Bamman2014DistributedLanguage}. That was introduced in evaluating language differences between different US states. We use the implementation provided by the authors.
\item \textit{Static-Word2vec} (\textit{SW2V}): a baseline adopted by \citeauthor{Yao2017DiscoveryLearning} and \citeauthor{Szymanski2017TemporalEmbeddings}. The embeddings are learned over all the diachronic corpus, ignoring the temporal slicing.
\end{itemize}
Note that in this task we also tested the model introduced by~\citeauthor{Rudolph2017DynamicEvolution} and we obtained results close to the baseline SW2V; these results have been confirmed by other authors in the literature~\citeauthor{barranco2018tracking}; Thus, we do not report the results for this model on the analogy task.

\subsubsection{Experiments on \ac{datas}}
The first setting involves all the presented models, trained on \ac{datas} and tested over \ac{tests}. The hyper-parameters reflect those of \citeauthor{Yao2017DiscoveryLearning}: we use small embeddings of size $50$, a window of $5$ words, $5$ negative samples and a small overall vocabulary of $21$k words with at least $200$ occurrences over the entire corpus.  Table \ref{test:set1:table_accuracysd} summarizes the results.

\begin{figure}[]
\centering
\includegraphics[width=0.75\columnwidth]{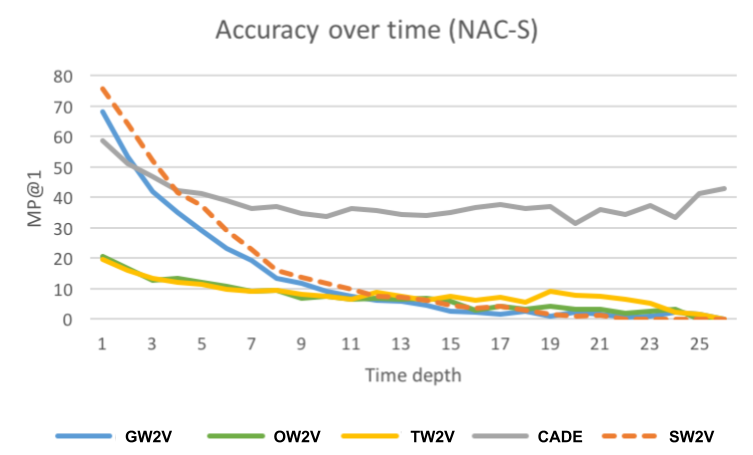}
\caption{Accuracy (MP@$1$) as function of time depth $\delta_t$ in \ac{tests}. Given an analogy $w_1 : w_2 = t_1 : t_2$, the time depth is plotted as $\delta_t=|t_1-t_2|$.}\label{test:set1:fig_accuracy}
\end{figure}

We can see that \ac{atmodel} outperforms the other models with respect to all the employed metrics. In particular, it performs better than DW2V, the second best model in the analogies, giving $7$\% more correct answers.
DW2V confirms its superiority with respect to the pairwise alignment methods, as in \citeauthor{Yao2017DiscoveryLearning}. Unfortunately, due to the lack of the answers set and the embeddings, we can not know how well it performs over static and dynamic analogies separately. TW2V and OW2V scored below the static baseline (as in \citeauthor{Yao2017DiscoveryLearning}), particularly on analogies with small time depth (see Figure \ref{test:set1:fig_accuracy}). In this setting, the pairwise alignment approach leads to huge disadvantages, probably due to data sparsity: the partitioning of the corpus produces tiny slices (around $3.5$k news articles) that are not sufficient to properly train the neural network; the poor quality of the embeddings affects the subsequent pairwise alignment. As expected, SW2V's accuracy on analogies drops sharply as time depth increases (Figure \ref{test:set1:fig_accuracy}). On the contrary, \ac{atmodel}, TW2V and OW2V maintain almost steady performances over different periods of time. GW2V does not answer correctly to almost any dynamic analogy. We conclude that GW2V alignment is not capable of capturing the semantic dynamism of words across time for the analogy task. For this reason, we do not employ it in our second setting.

\begin{table}[]
\small
\centering
\begin{tabular}{ccccccc}
\toprule  
Model & Set & MRR & MP1 & MP3 & MP5 & MP10\\ \midrule  
\multirow{2}{*}{SW2V} &Static&$\textbf{1}$&$\textbf{1}$&$\textbf{1}$&$\textbf{1}$&$\textbf{1}$\\ 
                        &Dynamic&$0.148$&$0.000$&$0.263$&$0.351$&$0.437$\\
                      &All&$0.375$&$0.266$&$0.459$&$0.524$&$0.587$\\ \midrule
\multirow{2}{*}{TW2V} &Static&$0.245$&$0.193$&$0.280$&$0.313$&$0.366$\\ 
                        &Dynamic&$0.106$&$0.069$&$0.123$&$0.156$&$0.205$\\
                      &All&$0.143$&$0.102$&$0.165$&$0.198$&$0.248$\\ \midrule
\multirow{2}{*}{OW2V} &Static&$0.265$&$0.202$&$0.299$&$0.348$&$0.415$\\ 
                        &Dynamic&$0.087$&$0.058$&$0.099$&$0.124$&$0.160$\\ 
                      &All&$0.135$&$0.096$&$0.153$&$0.183$&$0.228$\\ \midrule
\multirow{2}{*}{DW2V} &Static&$-$&$-$&$-$&$-$&$-$\\ 
                        &Dynamic&$-$&$-$&$-$&$-$&$-$\\ 
                      &All&$0.422$&$0.331$&$0.485$&$0.549$&$0.619$\\ \midrule
\multirow{2}{*}{GW2V} &Static&$0.857$&$0.819$&$0.888$&$0.909$&$0.931$\\ 
                        &Dynamic&$0.071$&$0.005$&$0.092$&$0.159$&$0.225$\\ 
                      &All&$0.280$&$0.222$&$0.305$&$0.359$&$0.435$\\ \midrule
\multirow{2}{*}{\textbf{\ac{atmodel}}} &Static&$0.720$&$0.668$&$0.763$&$0.787$&$0.813$\\ 
                        &Dynamic&$\textbf{0.394}$&$\textbf{0.308}$&$\textbf{0.451}$&$\textbf{0.508}$&$\textbf{0.571}$\\ 
                      &All&$\textbf{0.481}$&$\textbf{0.404}$&$\textbf{0.534}$&$\textbf{0.582}$&$\textbf{0.636}$\\ \bottomrule
\end{tabular}

\caption{MRR and MP for the subsets of static and dynamic analogies of \ac{tests}. We use MPK in place of MP@K.
DW2V results are taken from the original paper \cite{Yao2017DiscoveryLearning}.}
\label{test:set1:table_accuracysd}
\end{table}

\begin{figure}
\centering
\includegraphics[width=.75\columnwidth]{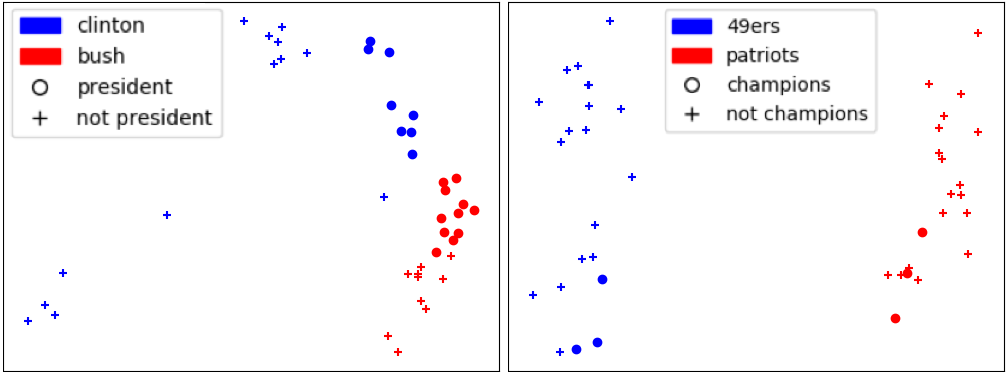}
\caption{2-dimensional PCA projection of the temporal embeddings of pairs of words from \textit{clinton}, \textit{bush} and \textit{49ers},  \textit{patriots}. The dot points highlight the temporal embeddings during their presidency or their winning years.}\label{qual:pairs_president}
\end{figure}

The comparison of the models' performances across the $25$ categories of analogies contained in \ac{tests} reveals new information: TW2V and OW2V's correct answers cover mainly $4$ categories, like \textit{President of the USA} and \textit{President of the Russian Federation}; \ac{atmodel} scores better across all the categories. Some categories are more difficult than others: even \ac{atmodel} scores nearly $0$\% in many categories, like \textit{Oscar Best Actor and Actress} and \textit{Prime Minister of India}. This discrepancy may be due to various reasons. First of all, some categories of words are more frequent than others in the corpus, so their embeddings are better trained. For example, \textit{obama} occurs $20,088$ times in \ac{datas}, whereas \textit{dicaprio} only $260$. As noted by \citeauthor{Yao2017DiscoveryLearning}, in the case of some categories of words, like presidents and mayors, the models are heavily assisted by the fact that they commonly appear in the context of a title (e.g. \textit{President Obama}, \textit{Mayor de Blasio}). For example in \ac{atmodel}, \textit{obama} during its presidency is always the nearest context embedding to the word \textit{president}. Lastly, as noted by \citeauthor{Szymanski2017TemporalEmbeddings}, some roles involved in the analogies only influence a small part of an entity's overall news coverage. We show that this is reflected in the vector space: as we can see in Figure \ref{qual:pairs_president}, presidents' embeddings almost cross each other during their presidency, because they share a lot of contexts; on the other hand, football teams' embeddings remain distant. This suggests that the capability of comparing word meanings across slices (contexts) may be affected by word frequency. We will further discuss this in the next sections.

\paragraph{Summary of the evaluation.}
\ac{atmodel} can effectively align temporal slices and its performances are better than the ones of the other models in most cases.

What makes each model different is how they estimate the change between slices. If the models overestimate change, they tend to get bad results on static; if they underestimate it, they tend to get bad results on dynamic analogies. \ac{atmodel} shows good capabilities in handling the balance in the estimation.

\subsubsection{Experiments on \ac{datab}}

This setting involves four models: SW2V, TW2V, OW2V, and \ac{atmodel}. The models are trained on \ac{datab} and tested over \ac{testb}. The parameters are similar to those of \citeauthor{Szymanski2017TemporalEmbeddings}: longer embeddings of size $100$, a window size of $5$, $5$ negative samples and a very large vocabulary of almost $200$k words with at least $5$ occurrences over the entire corpus. Table \ref{test:set2:table_accuracysd} summarizes the results.

\begin{table}[]
\centering
\small
\begin{tabular}{ccccccc}
\toprule  
Model & Set & MRR & MP$1$ & MP$3$ & MP$5$ & MP$10$\\
\midrule
\multirow{2}{*}{SW2V} &Static&$\textbf{1}$&$\textbf{1}$&$\textbf{1}$&$\textbf{1}$&$\textbf{1}$\\ 
                      &Dynamic&$0.102$&$0.000$&$0.149$&$0.259$&$0.326$\\ 
                        &All&$0.283$&$0.201$&$0.321$&$0.408$&$0.462$\\  \midrule
\multirow{2}{*}{TW2V} &Static&$0.842$&$0.805$&$0.869$&$0.890$&$0.915$\\  
                        &Dynamic&$0.343$&$0.287$&$0.377$&$0.414$&$0.467$\\ 
                      &All&$0.444$&$0.391$&$0.476$&$0.510$&$0.558$\\  \midrule
\multirow{2}{*}{OW2V} &Static&$0.857$&$0.824$&$0.876$&$0.903$&$0.926$\\ 
                        &Dynamic&$0.346$&$\textbf{0.290}$&$0.379$&$0.420$&$0.462$\\  
                      &All&$0.449$&$0.398$&$0.480$&$0.518$&$0.556$\\  \midrule
\multirow{2}{*}{\ac{atmodel}} &Static&$0.948$&$0.936$&$0.959$&$0.961$&$0.967$\\ 
                        &Dynamic&$\textbf{0.367}$&$0.287$&$\textbf{0.423}$&$\textbf{0.471}$&$\textbf{0.526}$\\ 
                      &All&$\textbf{0.484}$&$\textbf{0.418}$&$\textbf{0.531}$&$\textbf{0.570}$&$\textbf{0.615}$\\  \midrule
\end{tabular}
\caption{MRR and MP for the subsets of static and dynamic analogies of \ac{testb}. We use MPK in place of MP@K.}
\label{test:set2:table_accuracysd}
\end{table}

\ac{atmodel} still outperforms all the other models with respect to all the metrics, although its advantage is lower than in the previous setting. Table \ref{test:set2:table_accuracysd} shows that the advantage of \ac{atmodel} is limited to the static analogies. TW2V and OW2V score much better results than in the previous setting. This is due to the increased size of the input dataset which allows the training process to work well on individual slices of the corpus. 

\begin{figure}
\centering
\includegraphics[width=0.75\columnwidth]{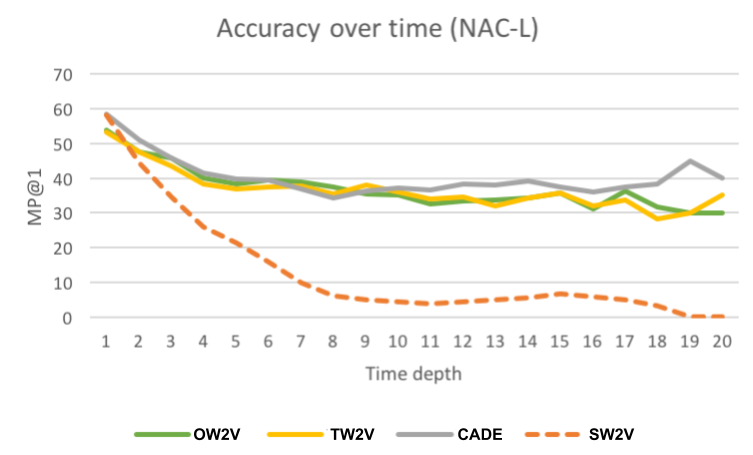}
\caption{Accuracy (MP@$1$) as function of time depth $\delta_t$ in \ac{testb}. Given an analogy $w_1 : w_2 = t_1 : t_2$, the time depth is plotted as $\delta_t=|t_1-t_2|$.}\label{test:set2:fig_accuracy}
\end{figure}

In Figure \ref{test:set2:fig_accuracy} we can see how, three temporal models behave similarly with respect to the time depth of the analogies. Performance stability is very different with respect from time depth. As expected, SW2V is the one that suffers the most on far-in-time-analogies.

\begin{figure}[]
\centering
\includegraphics[width=0.95\columnwidth]{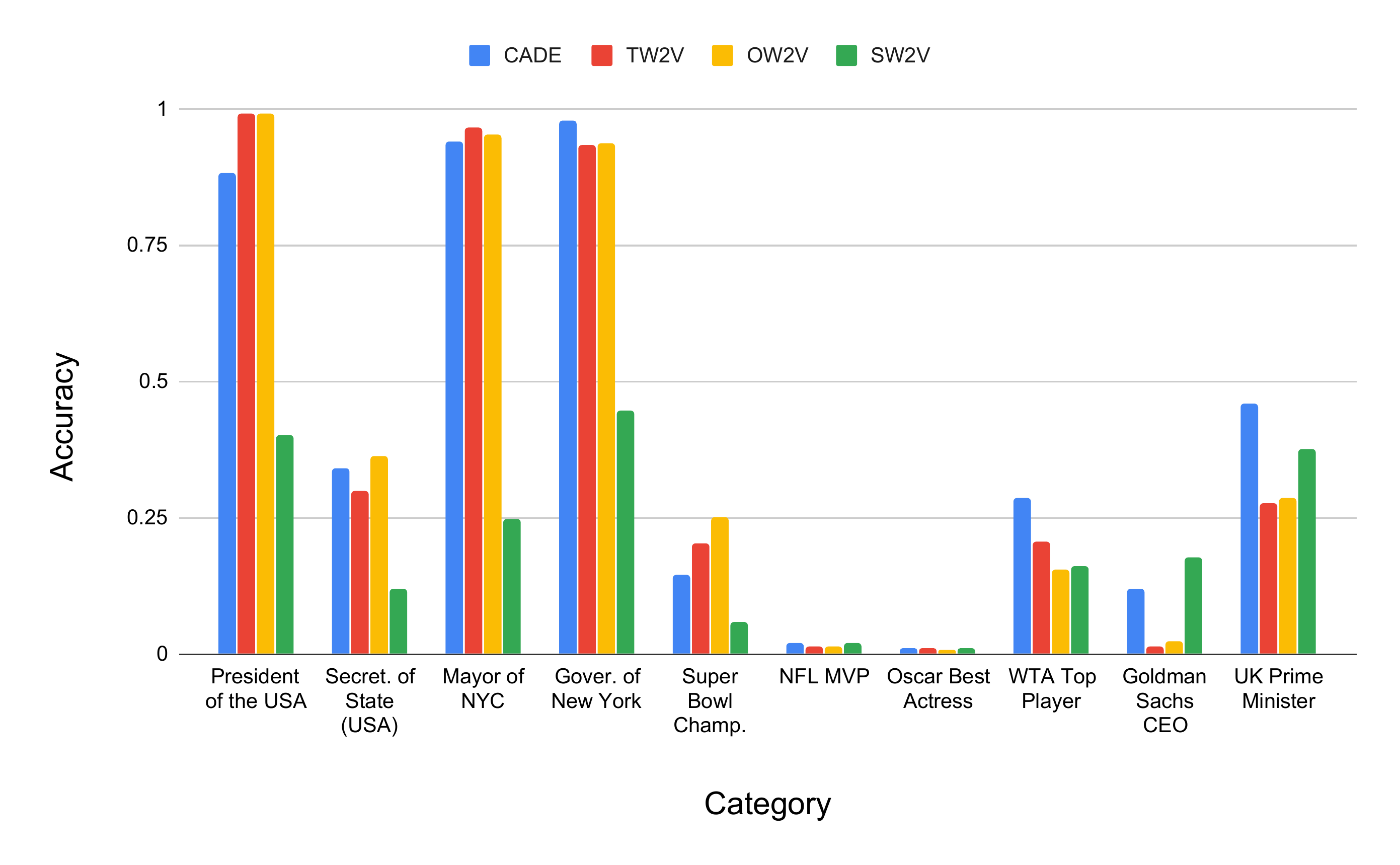}
\caption{Accuracy (MP@$1$) for the subsets of the analogy categories in \ac{testb}.}\label{fig:accuracy:categories}
\end{figure}

The comparison of the models' performances across the $10$ categories of analogies contained in \ac{testb} reveals more differences between them. The results in terms of MP@$1$ are summarized in Figure \ref{fig:accuracy:categories}. TW2V and OW2V outperform \ac{atmodel} in accuracy in two categories: \textit{President of the USA} and \textit{Super Bowl Champions}. In both cases, this is due to the major accuracy on dynamic analogies; for these categories, \ac{atmodel} is wrong because it gives static answers to dynamic analogies.
\ac{atmodel} significantly outperforms the other models in two categories: \textit{WTA Top-ranked Player} and \textit{Prime Minister of UK}. However in this case, \ac{atmodel} outperforms them both on dynamic and static analogies.

\subsection{Experiments with Held-Out Data}
In this section, we show the performance of the \ac{atmodel} on a held-out test task, in which we try to predict the \textit{slice} form which a held-out text comes from. We perform this test in two different ways. We tried to replicate the likelihood based experiments in~\citeauthor{Rudolph2017DynamicEvolution} and to further give confirmation about the performance of our model we also test the posterior probabilities using the framework described in~\citeauthor{taddy2015document}. 
Given a model, \citeauthor{Rudolph2017DynamicEvolution} assign a Bernoulli probability to the observed words in each held-out position: this metric is straightforward because it corresponds to the probability that appears in Equation~\ref{loss}. However, at the implementation level, this metric is highly affected by the magnitude of the vectors because it is based on the dot product of the vectors $\textbf{u}_k$ and $\textbf{c}_{\gamma(w_k)}$. In particular, \citeauthor{Rudolph2017DynamicEvolution} applied L2 regularization on the embeddings, which prioritize vectors with small magnitude.

This makes the comparison between models trained with different methods more difficult: regularization over the dot product can bias the comparison of the results. Furthermore, we claim that held-out likelihood is not enough to evaluate the quality of a \acf{twem}: a good temporal model should be able to extract features from each temporal slice that are discriminative and to improve the likelihood based on those features. To quantify this specific quality, we propose to adapt the task of document classification for the evaluation of \ac{twem}. We take advantage of the simple theoretical background and the easy implementation of the work of ~\citeauthor{taddy2015document}. We show that luckily this new metric is not affected by the different magnitude of the compared vectors. 

\begin{figure*}[h]
\centering
\includegraphics[width=0.95\textwidth]{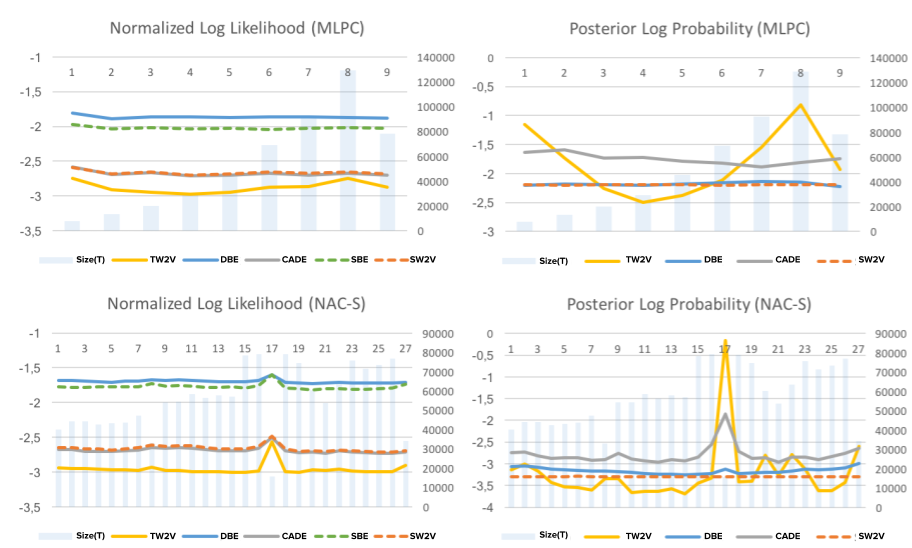}
\caption{$\mathcal{L}_{\mathcal{V}}^t$ and $\mathcal{P}_{\mathcal{V}}^t$ for each test slice $D^t$ and model $\mathcal{V}$. Blue bars represent the number of words in each slice.}
\label{img:likelihood:and:posteriors} 
\end{figure*}

\subsubsection{Datasets and Methodology}

We use two datasets for this task: the \ac{dataml} contains the full text from all the machine learning papers published on the ArXiv between $9$ years, from April 2007 to June 2015. The size of each slice is very small (less than $130,000$ words after pre-processing) and it increases over the years. \citeauthor{Rudolph2017DynamicEvolution} made \ac{dataml}  available online~\cite{Rudolph2017DynamicEvolution}: the text we obtain is already pre-processed, sub-sampled ($|V|=5,000$) and already split into training, validation and testing ($80\%$, $10\%$, $10\%$) to ease replication of the experiments. The data is shared in a computer-readable format without sentence boundaries: we convert it to plain text and we arbitrarily split it into 20-word sentences that are suited to cover our training procedure.
To make another comparison we also used the \ac{datas} dataset, which was described in previous sections and used to solve temporal word analogies. Consider that, compared to  \ac{dataml}, \ac{datas} has $\times3$ more slices and it has approximately $60,000$ words per slice, with a small exception due to the slice of the year 2006. To prepare this dataset for evaluation we use the same pre-processing script provided by \citeauthor{Rudolph2017DynamicEvolution} and divided the data training and testing ($|V|=21,000$). As before, details of these datasets are summarized in Table \ref{test:tabledata}.

We introduce new notation that will be helpful to understand this experiment, we will refer to $ \mathcal{V}=\{\mathcal{V}_{t \in 1 \cdots T}\} = \{ \langle \mathbf{C}^{t \in 1 \cdots T},\mathbf{U}^{t \in 1 \cdots T}\rangle \} $ as to the \ac{twem} taken in consideration and we will use $D^t$ to identify a temporal slice.

\noindent \textbf{Methodology A.}
We measure the held-out likelihood following a methodology similar to the one proposed by  \citeauthor{Rudolph2017DynamicEvolution}. Given a \ac{twem} $ \mathcal{V}=\{\mathcal{V}_{t \in 1 \cdots T}\} = \{ \langle \mathbf{C}^{t \in 1 \cdots T},\mathbf{U}^{t \in 1 \cdots T}\rangle \} $, we calculate the log-likelihood for the temporal testing slice $D^t=\langle w_1,\cdots,w_N \rangle$ as:
\begin{equation} \label{heldout_loglike}
\begin{split}
\log P_{\mathcal{V}_{t}}(D^t) & = \sum_{n=1}^{N} \log P_{\mathcal{V}_{t}}(w_n|\gamma(w_n))
\end{split}
\end{equation}
where the probability $\log P_{\mathcal{V}_{t}}(w_n|\gamma(w_n))$ is calculated based on Equation \ref{loss} using Negative Sampling and the vectors of $\mathbf{C}^t$ and $\mathbf{U}^t$. As \citeauthor{Rudolph2017DynamicEvolution}, we equally balance the contribution of the positive and negative samples. For each model $\mathcal{V}$, we report the value of the normalized log likelihood $\mathcal{L}^t$:
\begin{equation} \label{norm_log_like}
\mathcal{L}_{\mathcal{V}}^t = \frac{1}{N} \log P_{\mathcal{V}}(D^t)
\end{equation}
and its arithmetic mean $\mathcal{L}_{\mathcal{V}}$ over all the slices.

\noindent \textbf{Methodology B.}
We adapt the methodology of \citeauthor{taddy2015document} (\citeyear{taddy2015document}) to the evaluation of \ac{twem}. We calculate the posterior probability of assessing a temporal testing slice $D^t$ to the correct temporal class label $t$. In our setting, this corresponds to the probability that a model $\mathcal{V}$ predicts the year of the $t$-th slice given a held-out text that come from the same slice. We apply Bayes rules to calculate this probability: 
\begin{equation} P_{\mathcal{V}_t}(t|D^t)= \frac{P_{\mathcal{V}_t}(D^t)P(t)}{\sum_{k=1}^{T}P_{\mathcal{V}_k}(\mathcal{T}^t)P(k)}
\end{equation}
A good temporal model $\mathcal{V}=\{\mathcal{V}_{t \in 1 \cdots T}\} $ will assign a high likelihood to the slice $\mathcal{T}^t$ using the vectors of $\mathcal{V}_t$ and a relatively low likelihood using the vectors of $\mathcal{V}_{k\neq t}$.
We assume that the prior probability on class label $t$ is the same for each class, $P(t)=1/T$. We redefine the posterior likelihood as:
\begin{equation}
\mathcal{P}_\mathcal{V}^t=P_{\mathcal{V}_t}(t|\mathcal{T}^t)= \frac{1}{S} \sum_{s=1}^{S} \frac{P_{\mathcal{V}_t}({z}_s^t)P(t)}{\sum_{k=1}^{T} P_{\mathcal{V}_k}({z}_s^t)P(k)}
\end{equation}
where ${z}_s$ is the $s$-th sentence in $\mathcal{T}^t$ and $P_{\mathcal{V}_{t}}({z}_s)$ is calculated based on Equation \ref{heldout_loglike}. Please note that this metric is not affected by the magnitude of the vectors because is based on a ratio of probabilities. For each model $\mathcal{V}$, we report the value of the posterior log probability $\mathcal{P}_\mathcal{V}^t$ and its arithmetic mean $\mathcal{P}_\mathcal{V}$ over all the slices.

\subsubsection{Baselines}
We test five temporal embedding models for this setting: 

\begin{itemize}
    \item Our comparative framework \ac{atmodel}
    \item TW2V (a baseline used in previous experiments)
    \item SW2V (a baseline used in previous experiments)
    \item \textit{Dynamic Bernoulli Embeddings} (DBE) \cite{Rudolph2017DynamicEvolution} \item \textit{Static Bernoulli Embeddings} (SBE) \cite{rudolph2016exponential}.
\end{itemize}{}

Note that TW2V is equivalent to OW2V in this setting because we do not need to align vectors from different slices. DBE is the temporal extension of SBE, a probabilistic framework based on CBOW: it enforces similarity between consecutive word embeddings using a prior in the loss function, and specularly to \ac{atmodel}, it uses a unique representation of context embeddings for each word. We trained all the models on the temporal training slices $D^t$ using a CBOW architecture, a shared vocabulary and the same parameters, which are similar to \citeauthor{Rudolph2017DynamicEvolution}: learning rate $\eta=0.0025$, window of size $1$, embeddings of size $50$ and $10$ iterations ($5$ static and $5$ dynamic for \ac{atmodel}, $1$ static and $9$ dynamic for DBE as suggested by \citeauthor{Rudolph2017DynamicEvolution}). 
Following \citeauthor{Rudolph2017DynamicEvolution}, before the second phase of the training process of \ac{atmodel}, we initialize the temporal models with both the weight matrices $\mathbf{C}$ and $\mathbf{U}$ of the static model: we experimentally noted that this operation improves held-out performances but it negatively affects the analogy tests. 
We limit our study to small datasets and small embeddings due to the computational cost: DBE takes almost $6$ hours to train on \ac{datas} on a $16$-core CPU setting. DBE and SBE are implemented by the authors using \textit{tensorflow}, while all the other models are implemented in \textit{gensim}: to evaluate them, we convert them to \textit{gensim} models, extracting the matrices we need for comparison.

\subsubsection{Experimental Results}
Table \ref{test:heldout} shows the mean results of the two metrics for each model. In both settings, \ac{atmodel} obtain a likelihood almost equal to SW2V but a much better posterior probability than the baseline. This is remarkable considering that \ac{atmodel} optimizes the scoring function only on one weight matrix $\mathbf{C}^t$, keeping the matrix $\mathbf{U}^t$ frozen. With respect to TW2V, \ac{atmodel} has a better likelihood and its posterior probability is more stable across slices (Figure \ref{img:likelihood:and:posteriors}). 
The likelihood scores of DBE and SBE are highly influenced by the different magnitude of their vectors: we can quantify the contribution of the applied L2 regularization comparing the two static baseline SBE and SW2V. Differently from \ac{atmodel}, DBE slightly improves the likelihood with respect to its baseline. However, regarding the posterior probability, \ac{atmodel} outperforms DBE. Our experiments suggest an inverse correlation between the capability of generalization and the capability of extracting discriminative features from small diachronic datasets. Finally, experimental results show that \ac{atmodel} captures discriminative features from temporal slices without losing generalization power.

\begin{table}[h]
\small
\centering
\begin{tabular}{ccccccc}
\toprule
Dataset & M & SW2V & SBE & \ac{atmodel} & DBE & TW2V \\ \midrule
\multirow{2}{*}{\ac{dataml}} & $\mathcal{L}_{\mathcal{V}}$&-$2.67$&-$2.02$&-$2.68$&-$\textbf{1.86}$&-$2.88$\\ 
    
    &$\mathcal{P}_{\mathcal{V}}$&-$2.20$&-$2.20$&-$\textbf{1.75}$&-$2.18$&-$2.83$\\ 
                       \midrule
\multirow{2}{*}{\ac{datas}} &$\mathcal{L}_{\mathcal{V}}$&-$2.66$&-$1.77$&-$2.69$&-$\textbf{1.70}$&-$2.96$\\
                        &$\mathcal{P}_{\mathcal{V}}$&-$3.30$&-$3.30$&-$\textbf{2.80}$&-$3.16$&-$3.24$\\  \bottomrule
    
\end{tabular}
\caption{The arithmetic mean of the log likelihood $\mathcal{L}_{\mathcal{V}}$ and of the posterior log probability $\mathcal{P}_{\mathcal{V}}$ for each model $\mathcal{V}$. Based on the standard error on the validation set, all the reported results are significant.}
\label{test:heldout}
\end{table}

\subsection{Observations}

This experiment was meant to understand two things: how effective the alignment is and how general the features we learn are. 

To gather evidence for the first, we compared our model on a state-of-the-art task: temporal analogical reasoning. Results showed that our model reaches state-of-the-art performance and it is also stable: while other models tend to overestimate semantic shift, our method, that changes the representation only when contexts between slices change, is more careful in suggesting meaning shift.

For the second, we compared our model on another state-of-the-art task: held-out testing. We show that the model learns discriminative features and its results are comparable to the one obtained in the state-of-the-art.

\section{Generalization: Experiments on Language Localization and Topic-based Analyses}\label{sec:experiments:generalization}

\paragraph{Overview.}
To show how \ac{atmodel} can generalize the alignment between word vector spaces generated from different corpora we use a novel dataset built by scraping articles from news platforms. We will show how, with \ac{atmodel}, it is possible to compare two different corpora and show that it is possible to effectively discover orthographic differences between American English and British English.

\subsection{Quantitative Experiments on Language Localization with Newspaper Data}
In this section, we try to evaluate the generalization capabilities of our corpus-based comparative framework. To show how \ac{atmodel} can generalize the alignment starting from different corpora we use a novel dataset that contains text from newspapers.

\subsubsection{Datasets and Methodology}
\label{sec:experiments:generalization:datamethod}
We extracted articles from the New York Times and from The Guardian online platforms from the 9th of July 2019 to the 20th of September 2019. At the end of the process, we collected 14.480 articles from the New York Times, and 17.976 articles from The Guardian. We removed stop words from both the slices and brought the text to lowercase. Thus, in the context of our comparative framework, our collection $D$ contained two slices, one with the text from the New York Times and one with the text from The Guardian. We generated aligned embeddings with CADE that are 50 dimensional and are trained with a window size of 5.

As a baseline algorithm, we used MUSE~\cite{conneau2017word}, a multi-lingual alignment tool proposed by Facebook at ICLR 2018. This tool can align word embeddings from multiple languages without parallel corpora to support the alignment. We wanted to compare \ac{atmodel} to MUSE to see if a multi-lingual alignment tool is useful to identify meaning shifts.
In our case, we will use MUSE to try to align two different \textit{ways of writing} the the same language. To perform a fair comparison, we used the configuration of the algorithm suggested by the authors, and we trained the embeddings using the same procedure defined in the paper.

Note that it is difficult to define a dataset that contains pairs of these words because there are many implicit biases that we would have to make to generate a dataset like this one, for example ``which is the equivalent of \textit{jumper} in American?'' which might have multiple answers.

As a test dataset, we used a list of pairs of words that have minor spelling differences in British and American English\footnote{\url{http://www.tysto.com/uk-us-spelling-list.html}}: for example, British tend to use the form ``labelling'' while Americans use ``labeling''. From the corpus extracted online, we removed words that appear with a frequency lower than 20 and 50 times and those words that were not present in our corpora. We end up with two sets of 279 (\textbf{BAW1}) and 131 (\textbf{BAW2}) pairs of words. In a certain sense, these pairs of words can be interpreted as analogies between the British English space and the American English space.

\subsubsection{Experimental Results}
We compare \ac{atmodel} and MUSE on the same task: given a word in the English language and moving its vector representation, we wanted to find its equivalent in the American language (by looking at the neighborhood). We looked at the top-5 and top-10 neighbors, and we thus evaluated the HITS@5 and HITS@10 on both \textbf{BAW1} and \textbf{BAW2}. Results are visible in Tables~\ref{tab:spelling:results:one} and Tables~\ref{tab:spelling:results:two} and show that the \ac{atmodel} performs better than the competitor in this task. In general, while the alignment provided by MUSE is also good, it cannot use general contextual information that is what makes \ac{atmodel} more efficient (i.e., this is the effect of the compass). Another point is that with the increase of the frequency, both models become better with the mappings; these results confirm what was introduced in the experiment with temporal analogies: frequency is a key element to generate a good representation. Nevertheless, we underline that \ac{atmodel} is currently not able to align multi-language corpora that is the main area in which MUSE was proposed.

\begin{table}[h]
\centering
\begin{minipage}[b]{0.49\hsize}\centering
\begin{tabular}{ccc} \toprule
        Algorithm & HITS@5 & HITS@10 \\ \midrule
        \ac{atmodel} & \textbf{0.60}  & \textbf{0.64} \\
        MUSE & 0.40 & 0.56 \\ \bottomrule
    \end{tabular}
    \caption{British-American spelling test with words with frequency higher than 20 in the corpus (\textbf{BAW1})}
     \label{tab:spelling:results:one}
    \end{minipage}
    \hfill
    \begin{minipage}[b]{0.49\hsize}\centering
\begin{tabular}{ccc} \toprule
        Algorithm & HITS@5 & HITS@10 \\ \midrule
        \ac{atmodel} & \textbf{0.81}  & \textbf{0.86} \\
        MUSE & 0.51 & 0.60 \\ \bottomrule
    \end{tabular}
    \caption{British-American spelling test with words with frequency higher than 50 in the corpus (\textbf{BAW2})}
     \label{tab:spelling:results:two}
    \end{minipage}
\end{table}

\subsection{Qualitative Experiments on Language Localization with Newspaper Data}
We hereby show some examples of correspondence for British and American english that we are able to dected with CADE (e.g., \textit{biscuit}/\textit{cookie}, \textit{flat}/\textit{apartment}). 
We thus decided to collect some words of this kind and show which are their respective words in the other space; while this is not a quantitative experiment, it should give the reader the idea that the model is stable enough to be general and to map words that have the same meaning in the same space. For each word we also show the neighbourhood of that word in the mapped space; we do this last step to show that the matching words are not found in the neighbourhood (i.e., ``flat'' is not close to ``apartment'' in the NYT space). See Figure~\ref{tab:qualitative:examples:mapping} for some examples of differences between The New York Times and The Guardian that our model can be used to find.

\begin{table}[h]
    \centering
    \begin{tabular}{ccc} \toprule
            Mapping & Word & Top-5  \\ \midrule
            GUA $\rightarrow$ NYT & flat & `\textbf{apartment}', `walkup', `flat', `upstairs', `oneroom' \\
            NYT & flat & `sliding', `padding', `rough', `seams', `oneroom' \\ \midrule
            GUA $\rightarrow$ NYT & petrol & `\textbf{gasoline}', `idling', `trucks', `diesel', `suv' \\
            NYT & petrol & `lactic', `ricocheting', `nanoparticles', `quill', `squish' \\ \midrule
            GUA $\rightarrow$ NYT & garbage & `bins', `garbage', `litter', `\textbf{rubbish}', `bags' \\
            NYT  &  garbage& `trash', `cans', `bins', `piles', `bags' \\ \midrule
            NYT $\rightarrow$ GUA & candy & '\textbf{sweets}', 'chocolate', 'sip', `crisps', `jelly' \\
            GUA &  candy & `spears', `heavenly', `bud', `manger', 'jasmine' \\ \midrule
            NYT $\rightarrow$ GUA & gasoline & '\textbf{petrol}', 'diesel', 'fumes', `batteries', `plugin' \\
            GUA &  gasoline & `pellets', `dispose', `tubes', `microfibres', 'landfilled' \\ \midrule
            GUA $\rightarrow$ NYT & biscuits & `\textbf{cookies}', `chocolate', `pancakes', `bread', `noodles' \\
            NYT & biscuits & `honey', `vanilla', `spinach', `cinnamon', `coconut' \\ \bottomrule
        \end{tabular}
        \caption{Qualitative examples of mapping between the two spaces NYT and GUA}
         \label{tab:qualitative:examples:mapping}
\end{table}{}

\subsection{Qualitative Experiments on Topic-based Analyses with Reddit Boards Data}
Reddit is an online forum divided into \textit{boards}, main topics in which users can post related information. For example, the ``TwoXChromosomes'' describes itself has ``a subreddit for both serious and silly content, and intended for women's perspectives.``. Instead, the ``sports'' board is mainly used to share information about sports.

We use reddit data\footnote{\url{https://archive.org/details/2015_reddit_comments_corpus}} that was also used in a recent paper to drive domain-specific sentiment lexicons for different boards~\cite{hamilton2016inducing}. This corpus was generated by extracting data using APIs containing ~1.65 billion comments (of which 350,000 are not available as reported on the online page). The comments inside the complete dataset were produced from October of 2007 until May of 2015 but for these experiments only those from 2014 were considered.

We select some board pairs for which we expect to find cultural meaning shifts for the same words.

\paragraph{TwoXChromosomes vs. Sports.} In this subsection we show some examples related to the TwoXChromosomes board (TWX) and the Sports (SPO) board. These two boards capture context-specific meanings for the word ``period'': in TWX this word is frequently used to indicate female menstrual cycle. While in the SPO board this word is generally used to refer to a period of time during sports matches. We train \ac{atmodel} over this two slices and their concatenation as a compass and we show results in Table~\ref{tab:qualitative:examples:mapping:sport:twox}; as expected \ac{atmodel} can map the respective meaning in the correct positions.

\begin{table}[h]
    \centering
    \begin{tabular}{ccc} \toprule
            Mapping & Word & Top-5  \\ \midrule
            TWX & period & 'periods', 'spotting', 'cycle', `bleeding', `flow' \\
            SPO &  period & `periods', `duration', `stoppage', `half', 'longest' \\ \midrule
            TWX $\rightarrow$ SPO & period & `samples', `\textbf{blood}', `sprain', `stitches', `bruising' \\
            SPO $\rightarrow$ TWX & period & `\textbf{lifetime}', `span', `time', `continuation', `remainder' \\ \bottomrule
        \end{tabular}
        \caption{Qualitative examples of mapping between the two spaces SPO and TWX}
         \label{tab:qualitative:examples:mapping:sport:twox}
\end{table}{}

\paragraph{Science vs. Pokemon.} Another example we ran was on the Science (SCI) board and the Pokemon (POK) board. Pokemon have their own ``mythology'' and several meanings have a strong and different connotation from standard English. We use \ac{atmodel} to map the two spaces, results are visible in Table~\ref{tab:qualitative:examples:mapping:sci:poke}.
The word ``move'' is generally used to refer to a Pokemon attack (also referred to as technical move (tm)). The word ``ash'' in the Pokemon world is heavily influenced by the fact that the protagonist of the Pokemon tv-series is named ``Ash'. From the SCI space we can import the representation of the meaning of the word ``ash'' that is more related to a natural effect. Another interesting example is shown in the Table: ``Arceus'' is a Pokemon and in the POK space is close to other Pokemons. As soon as we move its vector to the SCI space we get in the vicinity of terms like ``gods'' and ``creator''. This happens because in Pokemon mythology the Pokemon Arceus is generally referred to as the creator of the Pokemon. Note that the term ``arceus'' is not present in the SCI corpus. \ac{atmodel} can map these semantic differences in language offering an interesting view on how to explain terms.

\begin{table}[h]
    \centering
    \begin{tabular}{ccc} \toprule
            Mapping & Word & Top-5  \\ \midrule
            SCI $\rightarrow$ POK & move & '\textbf{walk}', 'go', 'hop', `sit', `rotate' \\
            POK &  move & `tm', `moves', `moven', `superpower', 'attacks' \\ \midrule
            POK $\rightarrow$ SCI & arceus & '\textbf{gods}', 'sakes', 'worship', `creator', `god' \\
            POK  &  arceus & `mew', `deoxys', `giratina', `celebi', 'regigigas' \\ \midrule
            SCI $\rightarrow$ POK & ash & `lava', `moisture', `air', `ocean', `clouds' \\
            POK & ash & `brock', `misty', `serena', `giovanni', `gary' \\ \bottomrule
        \end{tabular}
        \caption{Qualitative examples of mapping between the two spaces}
         \label{tab:qualitative:examples:mapping:sci:poke}
\end{table}{}

\subsection{Observations}
This experiment was meant to show how it is possible to use \ac{atmodel} in a context that is different from the temporal one. We have shown that it is possible to align spaces generated from two corpora that are different (for example, British and American English). One key added value is that we are able to give explanations from some terms for which the core meaning strongly differs in the slices, has we have shown in the experiment with Arceus.

\section{Robustness: Experiments on Temporal Word Embeddings and Corrupted Corpora}\label{sec:experiments:robustness}

\paragraph{Overview.}

In this section, some of the assumptions that stand behind the model are explored. In general, \ac{atmodel} cannot be used in contexts where there is a low shared vocabulary between the slices. For example, using two different languages as slices generates two embeddings in which specific correspondences are not really informative. Hereby, we focus on the following objectives:

\begin{itemize}
    \item \textbf{Aligning the same corpus twice}: an important property to understand is how random is the effect of the training in \ac{atmodel}. What happens if we try to align two identical slices? our assumption is that our model works only if the alignment between two identical slices is close to perfect, meaning that the same words will have the same vectors. This differentiates our model from the standard word2vec, that generates different representations each time it is run on the same corpus.
    \item \textbf{Staticness}: another property to evaluate is how high is the tendency of the models to answer analogies with the same element they are given in input --- static analogies -- this experiment is devised to understand how static are the models. Temporal word embedding models are designed to represent the dynamics of word embeddings over time, as opposed to static models like word2vec. The less prone to change a temporal word embedding model is, the more \textit{static} its representation will be --- it will become more similar to a standard word2vec. we introduce the concept of staticness with respect to analogy solving, which shows quantitative results on how \ac{atmodel} can model change in the slices. 
    \item \textbf{Text Scrambling}: \ac{atmodel} bases itself on the presence of the same words in different corpora. With the following experiment we want to understand what happens when the semantics of the text is compromised by the injection of random noise: we randomly change words in the text with a varying probability $p$ and see how the similarity between words is affected.
    \item \textbf{Vocabulary Separation}: models based on the distributional hypothesis use contextual information to compute word vector representations. Specifically, for a single corpus language elements occurring in similar contexts are placed in similar positions in embedding space and vice-versa elements occurring in different contexts are placed far from one another; from this premise it is then clear that, if part of the vocabulary has no contextual overlap with the rest, it will have an isolated representation in embedding space, possibly even far from the rest of the vocabulary. By extending this reasoning to \ac{atmodel}, if two slices have little to no vocabulary overlap, the representation computed during the compass training on the corpora collection \textit{D} will diverge into two separated components since there is no shared context bringing their elements close together. An intuitive understanding of this phenomenon can be found through an analogy with spoken languages: speakers of two idioms with some common linguistic root (e.g., spanish and italian) can somehow understand each other due to the occasional shared word acting as a semantic “beacon” creating context from which to infer the meaning of a sentence. Languages that have no common ancestor, on the other hand, have no such link as there are no shared words to create common context (e.g., italian and japanese).

\end{itemize}{}

\subsection{Experiments on the Staticness of Temporal Word Embeddings}

\subsubsection{Dataset and Methodology}

In the experiment to evaluate the staticness, we use again \ac{datas} and \ac{tests}, the dataset and the testset for the temporal analogies.

The current literature does not offer a standardized metric to quantify how ``static'' the temporal word embeddings generated for different temporal slices are. We introduced a new metric to measure the tendency of a model to give the same answers of SW2V. Given a test set of analogies and the answers, the \ac{staticness} is measured as
$STAT= \frac{1}{N}\sum_{i=1}^{N} isstatic(i)$,
where the function $isstatic(i)$ is equal to $1$ if the answer given to the $i$-th analogy is static, $0$ otherwise. Staticness will be the ratio between the number of static analogies and the size of the test set: called the \textit{target staticness}.

We remark that a perfect model (we will refer to it as \textit{Target}) would have a $STAT=1.0$ over static analogies and a $STAT=0.0$ over dynamic ones;

The staticness of a model may affect its performances: a model that gives too many static answers will poorly perform over dynamic analogies and vice versa for dynamic answers. To verify this intuition we apply the metric in the two previous settings with the answers given by the models \ac{atmodel}, TW2V and SW2V; these models should illustrate the differences in performance with respect to the concept of staticness. We can apply \ac{staticness}  over subsets of the full test set, like static and dynamic analogies (note that over static analogies, staticness is equal to the accuracy).

\subsubsection{Experimental Results}

\begin{table}[!htb]
    \centering
    \begin{minipage}{.45\linewidth}
    \centering
        \begin{tabular}{cccc} \toprule
        Model & Subset & MP@$1$ & STAT\\ \midrule         
        SW2V&All&$0.2664$&$1.0000$\\ \midrule
        \multirow{3}{*}{TW2V} &All&$0.1019$&$0.0784$\\
                              &Static&$0.1933$&$0.1933$\\
                                &Dynamic&$0.0687$&$0.0367$\\  \midrule
        \multirow{3}{*}{\ac{atmodel}} &All&$0.4041$&$0.3024$\\
                                &Static&$0.6676$&$0.6676$\\
                              &Dynamic&$0.3082$&$0.1697$\\  \bottomrule
        \end{tabular}
        \caption{\ac{staticness}  and Accuracy (MP@$1$) for the analogies in \ac{tests}.}\label{qual:set1:table_staticness}
    \end{minipage}%
    \hfill
    \begin{minipage}{.45\linewidth}
    \centering
    \begin{tabular}{cccc} \toprule
    Model & Subset & MP@$1$ & STAT\\ \midrule          SW2V&All&$0.2014$&$1.0000$\\ \midrule
    \multirow{3}{*}{TW2V} &All&$0.3914$&$0.3057$\\
                          &Static&$0.8052$&$0.8052$\\
                            &Dynamic&$0.2867$&$0.1794$\\ \midrule
    \multirow{3}{*}{\ac{atmodel}} &All&$0.4183$&$0.4814$\\
                            &Static&$0.9362$&$0.9362$\\  
                          &Dynamic&$0.2873$&$0.3663$\\ \bottomrule
    \end{tabular}
    \caption{\ac{staticness} and Accuracy (MP@$1$) for the analogies in \ac{testb}.}\label{qual:set2:table_staticness}
    \end{minipage}
\end{table}

\begin{figure}[!tbp]
  \centering
  \begin{minipage}[b]{0.45\textwidth}
        \centering
        \includegraphics[width=1\textwidth]{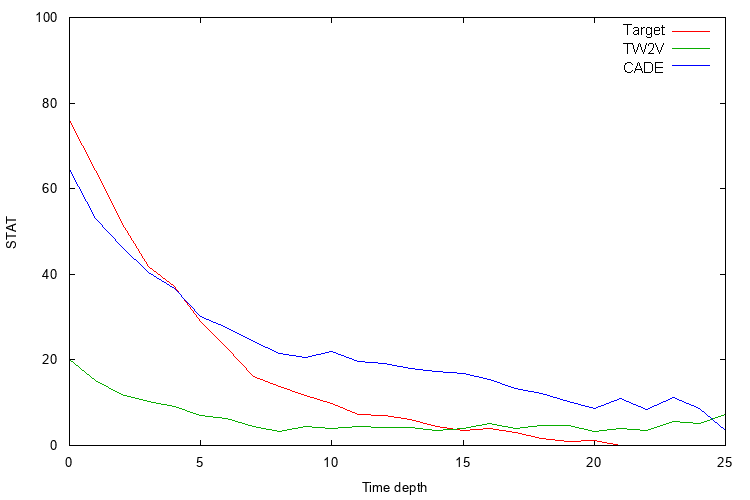}
        \caption{\ac{staticness} as function of time depth $\delta_t$ in \ac{tests}. Given an analogy $w_1 : w_2 = t_1 : t_2$, the time depth is plotted as $\delta_t=|t_1-t_2|-1$.}\label{qual:set1:static}
  \end{minipage}
  \hfill
  \begin{minipage}[b]{0.45\textwidth}
    \centering
    \includegraphics[width=1\textwidth]{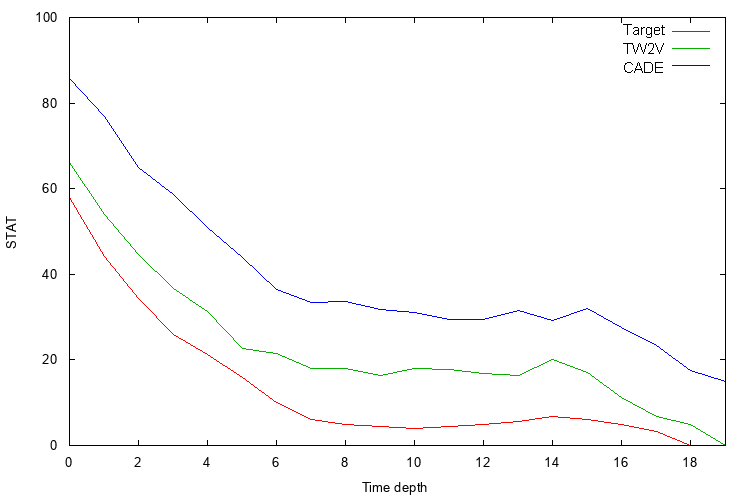}
    \caption{\ac{staticness} as function of time depth $\delta_t$ in \ac{testb}. Given an analogy $w_1 : w_2 = t_1 : t_2$, the time depth is plotted as $\delta_t=|t_1-t_2|-1$.}\label{qual:set2:static}
  \end{minipage}
\end{figure}

Table \ref{qual:set1:table_staticness} compares the results obtained by the two models with \ac{datas} and \ac{tests}. \ac{atmodel} goes slightly over the target value of $STAT=0.2664$; it gives static answers to $17$\% of the dynamic analogies and it gives dynamic answers to more than $30$\% of the static analogies of \ac{tests}. We can see it also in Figure \ref{qual:set1:static}, where it gives too many dynamic answers to analogies with small time depth, and too many static answers to analogies with greater time depth ($\delta_k>5$). However, TW2V performs badly in this test set and predicts dynamic answers more than $90$\% of the time. The poor performance of TW2V on dynamic analogies suggests that the temporal embeddings are highly dynamic due to their low quality and the insufficient training process.

Table \ref{qual:set2:table_staticness} reports the results relative to the big dataset settings, obtained with \ac{datab} and \ac{testb}. TW2V answers $30$\% of the time as SW2V, \ac{atmodel} almost half of the time, while the test set contains only $20$\% of static analogies. As a result, \ac{atmodel} scores better than TW2V on static analogies, but it scores the same as TW2V on dynamic analogies. This means that \ac{atmodel} gives less dynamic answers to dynamic analogies than TW2V but these dynamic answers are relatively more accurate. Moreover, \ac{atmodel} consistently gives more static answers then TW2V, independently from the time depth value (Figure \ref{qual:set1:static}). \ac{atmodel} generally represents the words giving more weight to the static aspect, i.e. the temporal embeddings of the same word are placed too close together. This leads to an advantage for \ac{atmodel} on static analogies but a disadvantage on dynamic ones. Moreover, we may expect that the static answer may ``hide'' sometimes the dynamic correct one. Indeed, incorrect static answers to dynamic analogies represent half the errors of \ac{atmodel} (twice those of TW2V); in $17$\% of these cases, the correct dynamic answer is just behind the incorrect static one (its rank is 2): it corresponds to a $5$\% drop in MP@$1$ on dynamic analogies. TW2V may have the opposite problem: its word representations are too dynamic; as a consequence, it gives less dynamic answers to static analogies than it should, but this may be an advantage with dynamic analogies.

The reason behind this discrepancy of staticness between TW2V and \ac{atmodel} may lie in the alignment strategy: global context approach may move the temporal embeddings of a word closer together because they share many words in their context. For example, all the temporal embeddings of the word \textit{bush} are close to the embedding of the word \textit{george}, because the two words appear nearby in all the time slices. However, the effects of the shared global context have to be further studied in more detail in future works.

\subsection{Experiments on Corrupted Corpora}

\subsubsection{Datasets and Methodology}
We laid out different experiments to evaluate the robustness of \ac{atmodel} to noise. We hereby summarize the dataset and the methodology of each experiment we will run.
\begin{itemize}
    \item \textbf{Alignment}: we evaluate the \textit{compass} using the \ac{atmodel} to align two slices that are composed of the same text. We created a collection with slices $D^1$ and $D^2$ where $D^2$ is an exact copy of $D^1$. We used \ac{atmodel} to align the two slices. For this experiment, \ac{atmodel} was trained 1000 times with a corpus of 5,000 articles that was copied to create the second corpus. The compass was the concatenation of the two corpora. Given the large number of times we trained this model to get a robust average, we decided to lower the dimensionality of the embeddings for this experiment to 20 to make computation faster.
    \item \textbf{Text Scrambling}: we used the same corpus used in the alignment setting and we corrupt part of the text by randomically replacing words inside the text. Thus corrupting the semantics that is used by distributional models to capture the meaning of words.
    \item \textbf{Vocabulary Overlap}: we evaluate the \ac{atmodel} model requirements for shared vocabulary as follows: from a single initial corpus (the collection of The Guardian news articles presented in Section~\ref{sec:experiments:generalization:datamethod}) a series of \textit{twin synthetic slice} pairs were generated, where an incremental fraction \textit{q} of vocabulary elements were “split” into different terms through the use of two suffixes, one for each twin slice. To give an example, for a given value of the separation rate \textit{q} a random sample of \textit{q} elements of the original corpus vocabulary are extracted; the twin slices are then generated by replacing all occurrences of the extracted terms with their suffixed counterparts (i.e., splitting the element \texttt{word} into \texttt{word\_A} and \texttt{word\_B}). The twin synthetic generated using the above procedure  are then aligned using \ac{atmodel} and the representations obtained evaluated, for each value of $q$, in two directions: \textit{same-word similarity} and \textit{correspondence matching}. For the \textit{same-word similarity} a set of random vocabulary elements was sampled\footnote{A different set was sampled for each value of $q$} and the cosine similarity between the respective representations in the twin slices was computed. The distribution of the resulting similarity scores were compared graphically. It should be noted that the evaluation was conducted separately for split and unsplit elements, providing nevertheless similar results. For the \textit{correspondence matching} evaluation, again a random set of vocabulary elements was sampled and the correspondence across slices was computed; a match was considered whenever  the nearest neighbour for the correspondence of a vocabulary element was its split twin (e.g., $\phi_{A \rightarrow B}(\texttt{word\_A}) \stackrel{?}{=} $ \texttt{word\_B}) or the word itself in case of unsplit elements. To further validate the results an additional corpus was created for each $q$ by shuffling one of the twin slices, thus removing the support for the shared context of each word (i.e., now each word will have random contexts). The resulting corpus was then aligned with the remaining unshuffled twin and the representations thus obtained used as baseline scenario, conducting the same evaluation procedure for each value of $q$.
\end{itemize}{}

\subsubsection{Experimental Results}

\paragraph{Aligning the same corpus twice.}
Using \ac{atmodel} to align the two identical slices, we obtained two embeddings for which the same words in the respective model have an average cosine similarity of 0.99 (computed on 30 different random samples of 1000 words).

Note that, words that occur less frequently tend to have a lower similarity due to a still existing intrinsic stochasticity in the network. Nevertheless, in general the alignment brings the two embeddings to be almost identical with respect to the cosine similarity.

\begin{figure}
\centering
\includegraphics[width=1\textwidth]{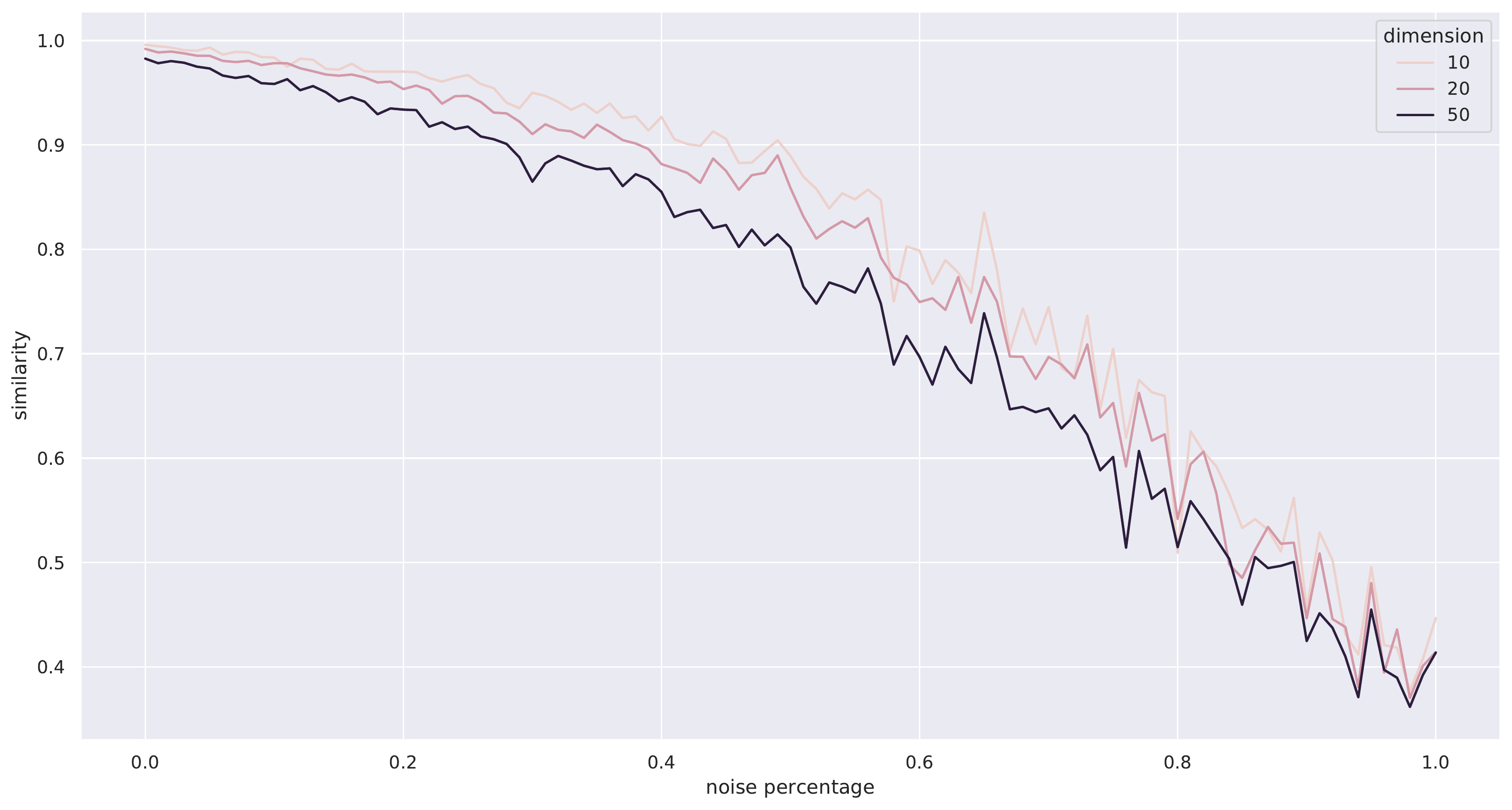}
\caption{Similarity between same words in \ac{atmodel} in the clean corpus and the noisy corpus while we vary the percentage of noise.}
\label{stability:noisy:effect:text}
\end{figure}

\paragraph{Text Scrambling.}
Starting from the input corpus (clean corpus) we generate a second corpus in which each word has a probability $p$ of being replaced with a different word randomly sampled from the vocabulary (noisy corpus). We use \ac{atmodel} to align the two corpora. At the end of the process 400 words from the vocabulary were sampled and the similarity between the aligned representations in the clean corpus and the noisy corpus was evaluated. We tested this methodology on embeddings of three different sizes: 10, 20 and 50. In Figure~\ref{stability:noisy:effect:text} we show the effect of this noising procedure on the words.

From the results it seems that the model can sustain a noise value around 20\% without losing too much in the similarity. After 20\% of noise the similarities start to drop quickly.

\paragraph{Vocabulary Overlap.}

\begin{figure}
\includegraphics[width=0.49\textwidth]{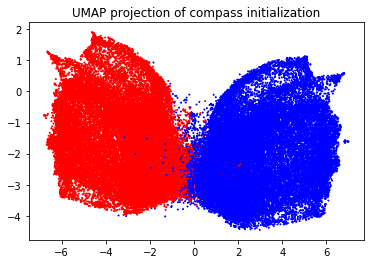}
\includegraphics[width=0.49\textwidth]{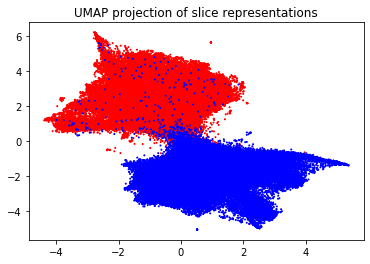}
\caption{UMAP projections of compass initialization vectors and aligned slices. Split vocabulary elements from each twin slice are coloured differently, showing the divergence in representation. }
\label{vocab:intersect:umap}
\end{figure}

Figure \ref{vocab:intersect:umap} shows the effect of vocabulary separation through a projection of the aligned representations of a pair of fully separated twin slices ($q=1.0$).

\begin{figure}
\includegraphics[width=0.49\textwidth]{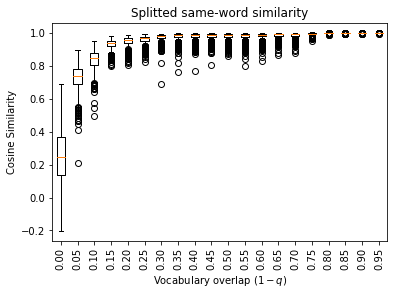}
\includegraphics[width=0.49\textwidth]{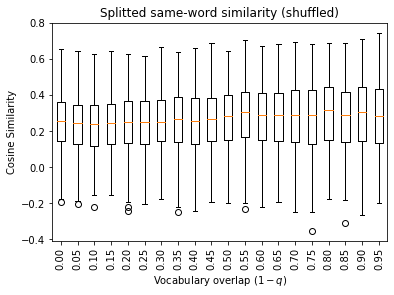}
\includegraphics[width=0.49\textwidth]{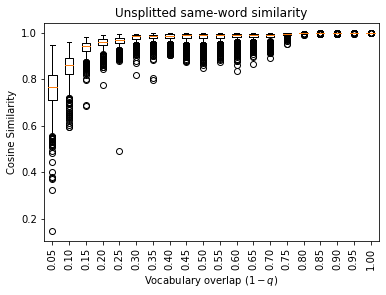}
\includegraphics[width=0.49\textwidth]{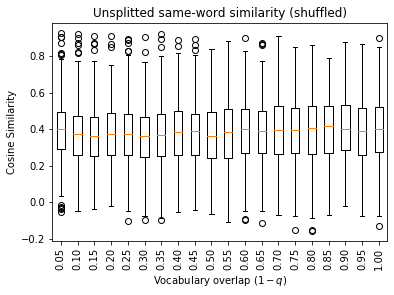}
\caption{Cosine similarity boxplots for same-word evaluation for different values of $q$ against the shuffled baseline. Each boxplot is a graphical representation of the similarity score distribution for a given value of $q$.}
\label{vocab:intersect:sameword}
\end{figure}

Figure \ref{vocab:intersect:sameword} shows how  \textit{same-word similarity} is strongly associated with vocabulary overlap $1-q$ for both split and unsplit vocabulary elements. The comparison to the shuffled baseline also suggests that the alignment does indeed capture semantic structures since the latter does not exhibit any dependence on vocabulary overlap.

\begin{figure}
\includegraphics[width=0.49\textwidth]{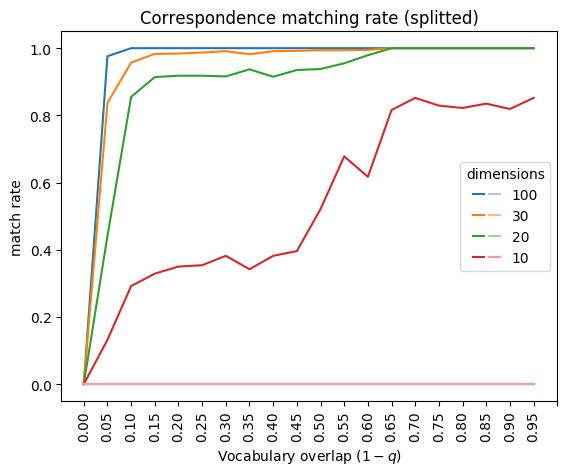}
\includegraphics[width=0.49\textwidth]{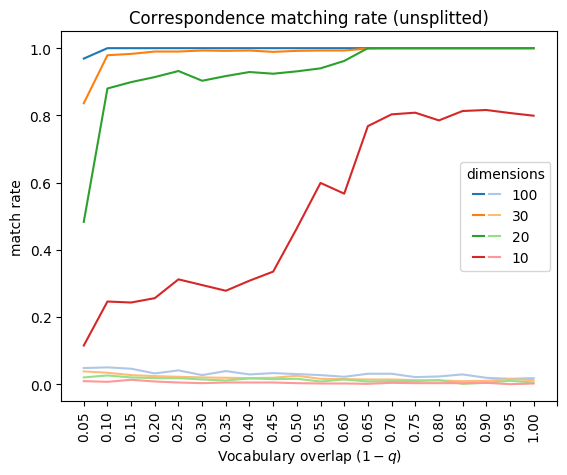}
\caption{Matching rate for correspondence evaluation for different values of $q$ against the shuffled baseline. }
\label{vocab:intersect:corrmatch}
\end{figure}

The results of the \textit{correspondence matching} evaluation are shown in Figure \ref{vocab:intersect:corrmatch}, where it is evident that model performance is still optimal even with just an overlap of 0.05 between vocabularies, before plummeting to zero matches for slices with no shared vocabulary. The shuffled baseline has instead an almost zero match rate, as is to be expected.

\subsection{Observations}
This experiment was meant to show how stable \ac{atmodel} is to noise. Our results show that our model is resistant to noise injection, but as soon as the semantic structure of the sentences is compromised, the model fails to provide a good alignment.

For the vocabulary overlap, the model produces good-quality alignments even at very high levels of separation. While the interpretation of this result must take into account that the semantic structure between twin slices was identical, it nevertheless suggests that the model can reliably operate even at low values of vocabulary overlap.


\section{Conclusions}\label{sec:conclusions}
In this paper, we have presented a general framework to support comparative cross-corpora language studies where the objective is to find semantic correspondences and differences between words based on their usage in different corpora. In particular, semantic comparisons are based on the comparisons among vector representations of words generated from different corpora. \ac{atmodel} is the core component of our framework, which provides an effective, general and robust method to align corpus-specific embeddings by using a \textit{compass} to share a coordinate system at training time. Such a general alignment method is needed as a prerequisite to support cross-corpora comparisons among word vectors. In addition to frame CASE within a comparative distributional framework, in the paper we have focused, in particular, on providing evidence about the effectiveness, generality and robustness of the proposed alignment method, discussing examples of comparisons in different domains along the evaluation process. We first explored the application to temporal word embeddings, showing that our model can effectively model change across corpora representing content produced in different time periods, a challenging application domain where comparative analysis with previous work is possible. Then, we have demonstrated that CADE is general enough to account for change in other domains. In particular, we have discussed experiments on language localization with British-English and American-English corpora and on topic-based analyses with different Reddit boards. Finally, we have discussed the robustness of the model and shed light on the conditions that make the alignment reliable. 

The method is based on word2vec and therefore limited in the respect that it considers only word-level embeddings and, therefore, embeddings where a word is associated to only one vector representing its core meaning. While cross-corpora semantic comparisons can track differences in meaning when words are used in different aggregated context, e.g., when writing about science vs. writing about pokemons, other solutions need and can be adopted to further mitigate the impact of polysemous words on such cross-corpora analyses. An example of such solution consists in pre-processig a collection with a word disambiguation technique, e.g., one focused on word sense disambiguation~\cite{iacobacci2015sensembed} or one focused on entity disambiguation~\cite{bianchi2018towards}. In this case, words are replaced by tokens, which can represent words or word sense and/or entity identifiers. Although a discussion of this problem is out of the scope of this paper, we believe that due to the abundance of word sense disambiguation~\cite{iacobacci2015sensembed} and named entity linking services~\cite{bianchi2018towards} a solution that combines these services and CADE can provide a quite straightforward, reasonable first solution to the polysemy problem in cross-corpora language studies. We have in fact carried out preliminary experiments with the combination of CADE and the distributional entity embeddings extracted from different corpora using the method discussed in~\cite{bianchi2018towards}, with encouraging preliminary results~\cite{bianchithesis}. However, a more thorough investigation of solutions to this problem will be the subject of future work. 

Another direction we plan to follow in future work is to combine the cross-corpora semantic comparison framework presented in this paper with language studies based on wordset associations, as the one developed by~\cite{caliskan2017semantics} using single-corpus embeddings. We believe that this direction is promising in tracking biases and differences across corpora.

\section{Acknowledgements}
This research has been supported in part by EU H2020 projects EW-Shopp - Grant n. 732590, and EuBusinessGraph - Grant n. 732003. We gratefully acknowledge the support of NVIDIA Corporation with the donation of the Titan Xp GPU used for this research.

\bibliography{sample}
\bibliographystyle{theapa}

\end{document}